\documentclass{article}

\usepackage{iclr2027_conference,times}

\usepackage{amsmath,amssymb,bm}
\usepackage{booktabs}
\usepackage{multirow}
\usepackage{graphicx}
\usepackage{microtype}
\usepackage{hyperref}
\hypersetup{hidelinks}   
\usepackage{url}
\usepackage{xcolor}
\usepackage{enumitem}
\usepackage[section,above,below]{placeins}   
\newcommand{\SO}{\mathrm{SO}}
\newcommand{\cF}{\mathcal{F}}
\newcommand{\cS}{\mathcal{S}}
\newcommand{\cA}{\mathcal{A}}
\newcommand{\cB}{\mathcal{B}}
\newcommand{\bx}{\bm{x}}
\newcommand{\bc}{\bm{c}}

\newcommand{\second}[1]{\underline{#1}}

\title{Low-Rank Clebsch--Gordan Coupling for \\Equivariant Spherical Forecasting}

\author{%
    Yaojun Li$^{*}$ \quad Zhixuan Zhao$^{*}$ \quad Dongzhe Zheng \quad Christine Allen-Blanchette\\
    Princeton University\\
    \texttt{\{yl3425, zz0487, dz5992, ca15\}@princeton.edu}\\
}

\iclrfinalcopy

\begin{document}

\maketitle
\fancyhead[L]{Preprint}

\begin{abstract}
Spherical-harmonic neural operators provide an efficient representation for global forecasting, but their discretized pointwise nonlinearities can break rotational equivariance through aliasing. Clebsch–Gordan (CG) tensor products provide exactly equivariant interactions between spherical-harmonic representations, but their computational cost grows rapidly with spectral bandwidth and has limited their use at forecasting scale. We introduce LCG-SFNO, which combines low-rank CG interactions within a controlled low-degree band with coefficient-space residual propagation and an exactly projected quadratic nonlinearity, preserving SO(3) equivariance throughout the learned operator. To make the CG interaction practical, we further develop SSHCG, a sparse GPU implementation with a factorized alternative for wider coupled bands. On spherical shallow-water dynamics and the ERA5-based WeatherBench benchmark, LCG-SFNO reduces rotational error to numerical precision—more than three orders of magnitude below SFNO on shallow water and five orders of magnitude below it on WeatherBench—while matching SFNO on shallow-water forecasting and substantially improving forecast skill relative to SFNO on WeatherBench. Matched ablations show that CG coupling accelerates convergence, while SSHCG keeps the coupling to about 10\% of a training step at the operating bandwidth, making exact equivariant coupling practical at forecasting scale.
\end{abstract}

\section{Introduction}
\label{sec:introduction}

Accurate global atmospheric forecasting supports applications in transportation,
energy, and disaster preparedness~\citep{lazo2009300,alley2019advances}. Numerical weather prediction provides
physically grounded forecasts but remains computationally expensive at high
resolution and for large ensembles~\citep{bauer2021digital,bauer2015quiet}. Data-driven models instead learn atmospheric evolution from reanalyses such as
ERA5~\citep{hersbach2020era5}, enabling substantially cheaper inference once
trained~\citep{pathak2022fourcastnet,lam2023graphcast,bi2023pangu,nguyen2023climax}.
Because atmospheric fields are defined globally, however, the prediction problem
is inherently spherical.

Spherical harmonics provide a natural basis for representing global atmospheric fields across spatial scales. Although this representation respects rotations, end-to-end equivariance requires every operation in the forecasting pipeline to preserve the same structure. Several forecasting architectures use grid- or position-dependent operations and therefore do not enforce rotational equivariance~\citep{pathak2022fourcastnet,tang2026gsno}. Even SFNO, despite operating in spherical harmonics, applies pointwise nonlinearities on a finite spatial grid. These nonlinearities can generate unresolved modes that alias into the retained spectrum, introducing orientation-dependent errors and allowing the predicted evolution to depend on the orientation used to represent the same physical state.

Clebsch--Gordan (CG) tensor products couple spherical-harmonic modes while preserving \(\SO(3)\) equivariance~\citep{kondor2018clebsch}. They therefore provide cross-degree interactions directly in coefficient space. However, dense CG evaluation becomes expensive at the bandwidths and channel widths required for forecasting. This motivates restricting the learned CG interaction to a controlled low-degree band while retaining full-band spectral processing elsewhere. 

Low-Rank Clebsch--Gordan Spherical Fourier Neural Operator
(LCG-SFNO) implements this design by combining full-band degree-wise spectral processing with low-rank CG coupling over a controlled low-degree band. A degree-wise pathway processes the full represented spectrum, and a CG correction introduces learned cross-degree interactions only within the coupled band. The hidden representation remains in spherical-harmonic coefficient space between blocks, and a quadratic nonlinearity with exact spectral projection prevents aliasing in the full-band nonlinear pathway from breaking equivariance. To make the CG interaction practical at forecasting scale, we also develop Sparse Spherical Harmonic Clebsch--Gordan (SSHCG), which provides sparse direct evaluation at moderate bandwidths and a factorized alternative for wider coupled bands.

Experiments on spherical shallow-water dynamics and WeatherBench show that
LCG-SFNO reduces end-to-end rotational error to numerical precision while
preserving forecast accuracy relative to SFNO. Matched ablations separate the
effects of the main architectural changes: CG coupling improves forecast error and
optimization, while dealiased nonlinear evaluation primarily reduces
end-to-end equivariance error. Runtime measurements show that the sparse direct
implementation is practical at the operating band, while the factorized route
provides better scaling for wider coupling.

We summarize our contributions as follows:
\begin{itemize}[leftmargin=6mm,topsep=0pt,itemsep=0pt]
    \item We introduce LCG-SFNO, an $\SO(3)$-equivariant spherical forecasting architecture combining full-band degree-wise spectral processing, low-rank CG cross-degree coupling, and a quadratic nonlinearity with exact spectral projection.
    \item We show on spherical shallow-water dynamics and WeatherBench that LCG-SFNO reduces end-to-end rotational error to numerical precision while preserving forecast accuracy relative to SFNO, and use matched ablations to separate the effects of CG coupling and dealiased nonlinear evaluation.
    \item We introduce SSHCG, which combines sparse direct CG kernels with a factorized evaluation for wider coupled bands, making CG interactions practical at forecasting scale.
\end{itemize}

\section{Related Work}
\label{sec:related-work}

\textbf{Data-driven global weather forecasting.}
Data-driven weather forecasting has progressed from grid-based convolutional and recurrent models
\citep{ronneberger2015unet,weyn2020improving,shi2015convlstm} to architectures designed for
global spatial interaction. Neural operators learn mappings between continuous fields
\citep{kovachki2023neural}, while Fourier-based models such as FNO and FourCastNet use spectral
convolution to exchange information globally
\citep{li2021fno,guibas2022afno,pathak2022fourcastnet}. More recent approaches increasingly
incorporate the geometry of the Earth, including spherical meshes in GraphCast
\citep{lam2023graphcast}, Earth-specific attention in Pangu-Weather
\citep{bi2023pangu}, and spherical harmonic representations in SFNO
\citep{bonev2023sfno,mahesh2024huge}. GSNO further introduces a position-dependent
Green's-function correction within a spherical neural operator \citep{tang2026gsno}.
These architectures increasingly account for global geometry, but they do not generally enforce
rotational equivariance throughout the forecasting operator. LCG-SFNO addresses this gap by
constructing both its spectral and nonlinear pathways to preserve rotational equivariance while
retaining the capacity required for global forecasting.

\textbf{Equivariant learning on the sphere.}
Equivariant neural networks constrain features to transform consistently under group actions.
Group-equivariant and spherical CNNs extend convolutional weight sharing to signals on $S^2$ and
$\SO(3)$ \citep{cohen2016group,cohen2018spherical,esteves2018learning,ocampo2023disco}, while
steerable and tensor-field networks organize features into irreducible representations of the rotation
group \citep{thomas2018tensor,weiler2018steerable,fuchs2020se3,geiger2022e3nn}. Related
tensor-product constructions are widely used in equivariant molecular modeling
\citep{batzner2022nequip,batatia2022mace}, while other approaches relax exact symmetry
\citep{wang2022approximately} or use alternative geometric representations
\citep{brandstetter2023clifford,zhdanov2024clifford}. Nonlinear interactions between irreducible
representations can be expressed exactly using Clebsch--Gordan (CG) tensor products
\citep{kondor2018clebsch}, but their cost grows rapidly with spectral bandwidth. LCG-SFNO addresses
this limitation by restricting low-rank CG interactions to a controlled spectral band while using
exactly projected polynomial nonlinearities to preserve equivariance through the remaining nonlinear
pathway.

\textbf{Efficient Clebsch--Gordan interactions.}
The cost of equivariant tensor products has motivated several more efficient formulations.
$\SO(3)$-to-$\SO(2)$ reductions lower the complexity of the underlying group operations
\citep{passaro2023reducing}, while Gaunt tensor products exploit the relation between
spherical-harmonic products and physical-space multiplication \citep{luo2024gaunt}; general-purpose
libraries such as \texttt{e3nn} provide optimized implementations of Clebsch--Gordan interactions
\citep{geiger2022e3nn}. These approaches reduce the cost of equivariant coupling, but broad-band
tensor products remain expensive at the scales considered here. SSHCG addresses this computational
bottleneck with sparse direct kernels for moderate bandwidths and a factorized evaluation for wider
bands, making explicit Clebsch--Gordan interactions practical within LCG-SFNO.

\section{Background}
\label{sec:background}

We review spherical harmonics and their transformation under rotations, together with equivariant nonlinear interactions in coefficient space. Appendix~\ref{app:background} provides additional details on the representation theory and quadrature conditions for exact spectral projection.

\textbf{Spherical harmonics and rotations.}
We expand a field $u$ on the sphere in complex spherical harmonics through
degree $L$,
\begin{equation}
u(\theta,\varphi)
=
\sum\nolimits_{\ell=0}^{L}
\sum\nolimits_{m=-\ell}^{\ell}
u_{\ell m}Y_{\ell m}(\theta,\varphi),
\qquad
u_{\ell m}
=
\int_{S^2}
u(\theta,\varphi)
\overline{Y_{\ell m}(\theta,\varphi)}
\,\mathrm{d}\Omega .
\label{eq:sh-expansion}
\end{equation}
Lower degrees represent larger angular scales, while higher degrees resolve
finer structure. We denote spherical-harmonic transform and its inverse by
$\operatorname{SFT}$ and $\operatorname{ISFT}$, respectively.

A rotation $R\in\mathrm{SO}(3)$ acts independently within each harmonic degree,
\begin{equation}
(Ru)_{\ell m}
=
\sum\nolimits_{m'=-\ell}^{\ell}
D^\ell_{m m'}(R)\,
u_{\ell m'},
\label{eq:wigner}
\end{equation}
where $D^\ell(R)$ is the degree-$\ell$ Wigner $D$-matrix
~\citep{edmonds1996angular}. The coefficients at each degree therefore transform
according to an irreducible representation of $\mathrm{SO}(3)$.

For a forecasting operator $\mathcal{G}$ with state $\mathbf{x}$ and spatial
conditioning fields $\mathbf{c}$, rotational equivariance requires
\begin{equation}
\mathcal{G}(R\mathbf{x},R\mathbf{c})
=
R\,\mathcal{G}(\mathbf{x},\mathbf{c}).
\label{eq:conditioned-equivariance}
\end{equation}
Since rotations mix orders $m$ only within each degree, a linear spectral map
whose channel mixing is shared across $m$ commutes with $D^\ell(R)$ and is
therefore equivariant. This degree-wise structure provides the equivariant
spectral processing used in Section~\ref{sec:spectral}.

\textbf{Equivariant nonlinearities.}
Nonlinear operations couple spherical-harmonic modes and can generate degrees
beyond the represented band. For example, the product of fields band-limited at
degrees $L_1$ and $L_2$ contains harmonics through degree $L_1+L_2$.
Consequently, nonlinear operations evaluated through a spatial discretization
require sufficient quadrature to recover their projection onto the retained
spectral band without aliasing.

Clebsch--Gordan (CG) products provide an equivariant construction of such
interactions directly in coefficient space. For coefficient sets $p$ and $q$,
\begin{equation}
z_{\ell m}
=
\sum\nolimits_{\ell_1,\ell_2}
\sum\nolimits_{m_1=-\ell_1}^{\ell_1}
\sum\nolimits_{m_2=-\ell_2}^{\ell_2}
C^{\ell m}_{\ell_1 m_1,\ell_2 m_2}
\,p_{\ell_1 m_1}
\,q_{\ell_2 m_2},
\label{eq:cg-product}
\end{equation}
where nonzero couplings satisfy
\begin{equation}
|\ell_1-\ell_2|
\leq
\ell
\leq
\ell_1+\ell_2 .
\label{eq:cg-selection}
\end{equation}
For each admissible degree triple, the CG coefficients define an intertwiner
between the corresponding $\mathrm{SO}(3)$ representations, so the resulting
coupling is equivariant
~\citep{thomas2018tensor,kondor2018clebsch,geiger2022e3nn}.

Direct angular contraction over degrees through $L_c$ scales as
$O(L_c^5)$. Polynomial nonlinearities provide another route to nonlinear
spectral interactions: because their output bandwidth is bounded, projection
onto a retained harmonic band can be evaluated exactly with sufficiently
accurate quadrature.

\section{Method}
\label{sec:method}
\begin{figure}[t]
\centering
\includegraphics[width=1.0\linewidth]{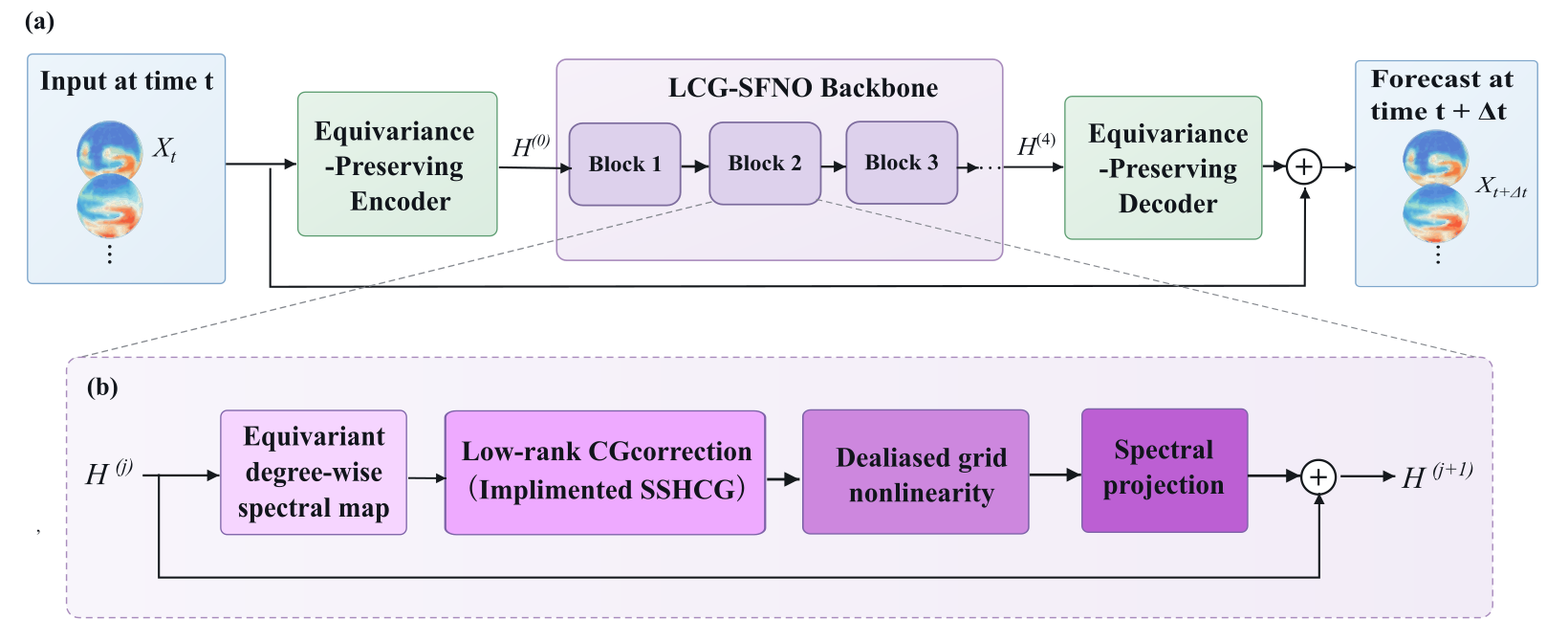}
\caption{
\textbf{Overview of LCG-SFNO.}
(a) Input fields are encoded into spherical-harmonic coefficients, processed by
coefficient-space residual blocks, and decoded to produce the forecast.
(b) Each block combines an equivariant degree-wise spectral map, a low-rank
Clebsch--Gordan correction evaluated with SSHCG, and a quadratic nonlinearity
with exact spectral projection. The residual state remains in coefficient space
between blocks.}
\label{fig:architecture}
\end{figure}

We introduce LCG-SFNO, an $\SO(3)$-equivariant spherical forecasting
architecture combining full-band degree-wise spectral processing, low-degree
Clebsch--Gordan coupling, and a quadratic nonlinearity with exact spectral
projection. Together, these operations preserve rotational equivariance
throughout the forecasting operator. We additionally develop SSHCG for
efficient evaluation of the CG interaction.

\subsection{Model Overview}
\label{sec:overview}

Given an input state $\mathbf{x}_t$ defined on the sphere and spatial
conditioning fields $\mathbf{c}$, an encoder $E$ maps the inputs to a latent
field, which is transformed to spherical-harmonic coefficients
$H^{(0)}\in\mathbb{C}^{C\times(L+1)^2}$, where \(C\) is the latent channel width and \(L\) is the maximum represented harmonic degree. The coefficients are then processed by
a stack of $J$ coefficient-space residual blocks and decoded to obtain the
predicted state:
\begin{equation}
  H^{(0)}
  =
  \operatorname{SFT}\!\left[E(\mathbf{x}_t,\mathbf{c})\right],
  \qquad
  H^{(j+1)}
  =
  H^{(j)}+\Delta_j\!\left(H^{(j)}\right),
  \qquad
  \widehat{\mathbf{x}}_{t+\Delta t}
  =
  D\!\left(H^{(J)},\mathbf{x}_t\right)+S\mathbf{x}_t .
  \label{eq:pipeline}
\end{equation}
Here, \(\Delta_j\) is the residual update of the \(j\)th block, \(D\) is the decoder, and \(S\) is a pointwise linear map from the input state to the output space.

Each residual update combines a full-band degree-wise spectral map, a low-rank
CG correction over a low-degree band, and a quadratic nonlinearity with exact
spectral projection. The encoder and decoder use analogous equivariance-preserving
spectral operations; implementation details are given in Appendix~\ref{app:codec}.

\subsection{Equivariant Spectral Processing}
\label{sec:spectral}

LCG-SFNO mixes latent channels independently within each spherical-harmonic
degree. For coefficients $H_{b\ell m}$, we define
\begin{equation}
    [WH]_{a\ell m}
    =
    \sum\nolimits_b W_{\ell ab} H_{b\ell m},
    \label{eq:degree-linear}
\end{equation}
where $W_\ell$ is a learned channel-mixing matrix shared across all orders $m$
at degree $\ell$. A rotation acts on the order index through $D^\ell(R)$,
whereas $W_\ell$ acts only on the channel index. The two operations therefore
commute, making the spectral map $\SO(3)$-equivariant.

We apply this map at every represented degree, providing full-band spectral
processing. Because different degrees are processed independently, however,
the map does not exchange information across harmonic degrees. We introduce
this interaction through Clebsch--Gordan coupling in the next subsection.
Additional details on the spectral representation and weight constraints are
given in Appendix~\ref{app:spherical}.

\subsection{Low-Rank Clebsch--Gordan Coupling}
\label{sec:cg}

We introduce cross-degree interactions through a low-rank Clebsch--Gordan (CG)
correction. We restrict the CG inputs to degrees $\ell_1,\ell_2\leq L_c$ and retain outputs
through degree $L_o$. 
For each degree $\ell$, two learned projections reduce the $C$
hidden channels to $r$ streams,
\begin{equation}
p^{(s)}_{\ell m}
=
\sum\nolimits_b P_{\ell sb} H_{b\ell m},
\qquad
q^{(s)}_{\ell m}
=
\sum\nolimits_b Q_{\ell sb} H_{b\ell m},
\qquad
s=1,\ldots,r .
\label{eq:cg-projections}
\end{equation}
The rank-$r$ factorization reduces the channel dimension before the CG product.
For each stream, we then couple harmonic degrees through
\begin{equation}
z^{(s)}_{\ell m}
=
\sum\nolimits_{\ell_1,\ell_2\leq L_c}
\gamma^{(s)}_{\ell_1\ell_2\ell}
\sum\nolimits_{m_1,m_2}
C^{\ell m}_{\ell_1m_1,\ell_2m_2}
p^{(s)}_{\ell_1m_1}
q^{(s)}_{\ell_2m_2},
\qquad
\ell\leq L_o ,
\label{eq:cg-product-lowrank}
\end{equation}
where $\gamma^{(s)}_{\ell_1\ell_2\ell}$ is a learned weight for each admissible
degree path. The resulting features are projected back to the model channels,
\begin{equation}
\Delta H^{\mathrm{CG}}_{a\ell m}
=
\alpha_\ell
\sum\nolimits_{s=1}^{r}
O_{\ell as} z^{(s)}_{\ell m},
\qquad
\ell\leq L_o ,
\label{eq:cg-coupling}
\end{equation}
where $\alpha_\ell$ is a learned degree-wise scale initialized at zero.
Only degree triples satisfying
$|\ell_1-\ell_2|\leq\ell\leq\ell_1+\ell_2$
contribute.

The CG coefficients are fixed, while $P$, $Q$, $O$, and the degree-path
weights $\gamma$ are learned. These learned transformations are shared across
orders $m$ and therefore do not alter the action of rotations within each
degree. Together with the intertwining property of the CG tensor product, the
complete correction is $\SO(3)$-equivariant.

Since the product of two fields band-limited at $L_c$
contains harmonics only through degree $2L_c$, choosing $L_o=2L_c$ retains the
complete CG correction generated by the coupled band. Degrees above $L_o$
remain in the full-band spectral pathway. Appendix~\ref{app:band} examines the
choice of coupled bandwidth.

\subsection{Dealiased Spectral Nonlinearity}
\label{sec:dealiasing}

The degree-wise spectral map and CG correction are equivariant in coefficient
space, but nonlinear processing across the full represented band requires
transforming to a spatial grid. A pointwise nonlinearity can generate harmonics
beyond the represented degree $L$; if these modes are not sufficiently resolved
by the quadrature grid, they alias into the retained coefficients and break
equivariance. We avoid this aliasing by using a quadratic activation whose
projection onto the retained spectral band can be evaluated exactly.

Specifically, we use
\(
\sigma(x)=\tfrac{x}{2}+\tfrac{x^2}{4}
\)
and evaluate it on an oversampled Legendre--Gauss grid. Suppressing the block
index $j$, the residual update is
\begin{equation}
\Delta(H)
=
N_{\mathrm{spec}}
\left[
\Pi_L \operatorname{SFT}_f
\left(
B\,\sigma\!\left(
A\,\operatorname{ISFT}_f
\left(
WH+\Delta H^{\mathrm{CG}}
\right)
\right)
\right)
\right],
\label{eq:grid-update}
\end{equation}
where $\operatorname{ISFT}_f$ and $\operatorname{SFT}_f$ denote the inverse
and forward spherical Fourier transforms on a grid oversampled by a factor $f$,
and $\Pi_L$ projects onto degrees $\ell\leq L$. The maps $A$ and $B$ mix
channels before and after the activation, while $N_{\mathrm{spec}}$ denotes the
equivariance-preserving spectral normalization used in each block.

For an input band-limited at degree $L$, the quadratic activation generates
harmonics only through degree $2L$. Exact evaluation of its projection onto
degrees $\ell\leq L$ therefore requires only sufficient quadrature, rather than
explicit representation of all degree-$2L$ modes. By the classical
three-halves rule, an oversampling factor $f=1.5$ is sufficient for this
projection. Higher-degree content consequently does not alias into the retained
coefficients, and the nonlinear update remains $\SO(3)$-equivariant up to
numerical precision. The quadrature derivation, spectral normalization, and
comparison with non-polynomial activations are given in
Appendix~\ref{app:dealiasing}.

\subsection{Efficient Clebsch--Gordan Evaluation}
\label{sec:acceleration}

Direct evaluation of the CG interaction requires summing over all admissible
degree paths and angular couplings. For a coupled band extending through degree
$L_c$, the number of admissible degree triples scales as $O(L_c^3)$, while
direct angular contraction scales as $O(BCL_c^5)$, where $B$ is the batch size
and $C$ is the number of coupled channels. To reduce this cost, we develop
SSHCG, which provides complementary direct and factorized evaluation strategies.

For moderate coupled bandwidths, SSHCG uses a sparse direct implementation that
traverses only nonzero CG couplings and avoids dense angular tensors and
per-path intermediates. This preserves an independent learned weight for every
admissible degree path while retaining the $O(L_c^5)$ bandwidth scaling of
direct contraction.

For wider bands, SSHCG imposes a rank-$R$ canonical polyadic factorization on
the degree-path weights. The resulting interaction can be evaluated through
spherical synthesis, bilinear operations on the sphere, and harmonic analysis,
reducing the complexity to $O(BCRL_c^3)$. The factorized route therefore trades
unrestricted degree-path weights for more favorable bandwidth scaling.
Implementation details, derivations, and runtime comparisons are given in
Appendix~\ref{app:sshcg-direct}--\ref{app:speed}.

\section{Experiments}
\label{sec:experiments}
\begin{table}[t]
  \caption{\textbf{Rotational equivariance.} Relative equivariance error under joint rotations of the state and auxiliary spatial fields, averaged over eight rotations and sixteen test states. LCG-SFNO reduces SFNO's error by over three orders of magnitude on SSWE and approximately five on WeatherBench, reaching numerical precision. Best/second-best values are bold/underlined.}
  \label{tab:equiv}
  \vspace{1mm}
  \centering
  \small
  \setlength{\tabcolsep}{5pt}
  \begin{tabular}{lcccc}
    \toprule
    & Shallow water & \multicolumn{3}{c}{WeatherBench $5.625^\circ$} \\
    \cmidrule(lr){3-5}
    Model & $256\times512$ & 1 day & 3 days & 5 days \\
    \midrule
    ClimaX            & $8.5\times10^{-1}$ & $7.9\times10^{-1}$ & $1.0$ & $1.0$ \\
    FourCastNet       & $6.9\times10^{-1}$ & $7.6\times10^{-1}$ & $8.3\times10^{-1}$ & $8.3\times10^{-1}$ \\
    SHNet (GSNO)      & $1.1\times10^{-1}$ & $5.6\times10^{-1}$ & $7.2\times10^{-1}$ & $6.7\times10^{-1}$ \\
    SFNO              & \second{$9.5\times10^{-4}$} & \second{$9.5\times10^{-2}$} & \second{$1.5\times10^{-1}$} & \second{$1.1\times10^{-1}$} \\
    \textbf{LCG-SFNO (ours)}
                       & $\mathbf{8.2\times10^{-7}}$
                       & $\mathbf{6.6\times10^{-7}}$
                       & $\mathbf{7.1\times10^{-7}}$
                       & $\mathbf{7.2\times10^{-7}}$ \\
    \bottomrule
  \end{tabular}
\end{table}
\begin{figure}[b]
\vspace{-3mm}
\centering
\includegraphics[width=0.78\linewidth]{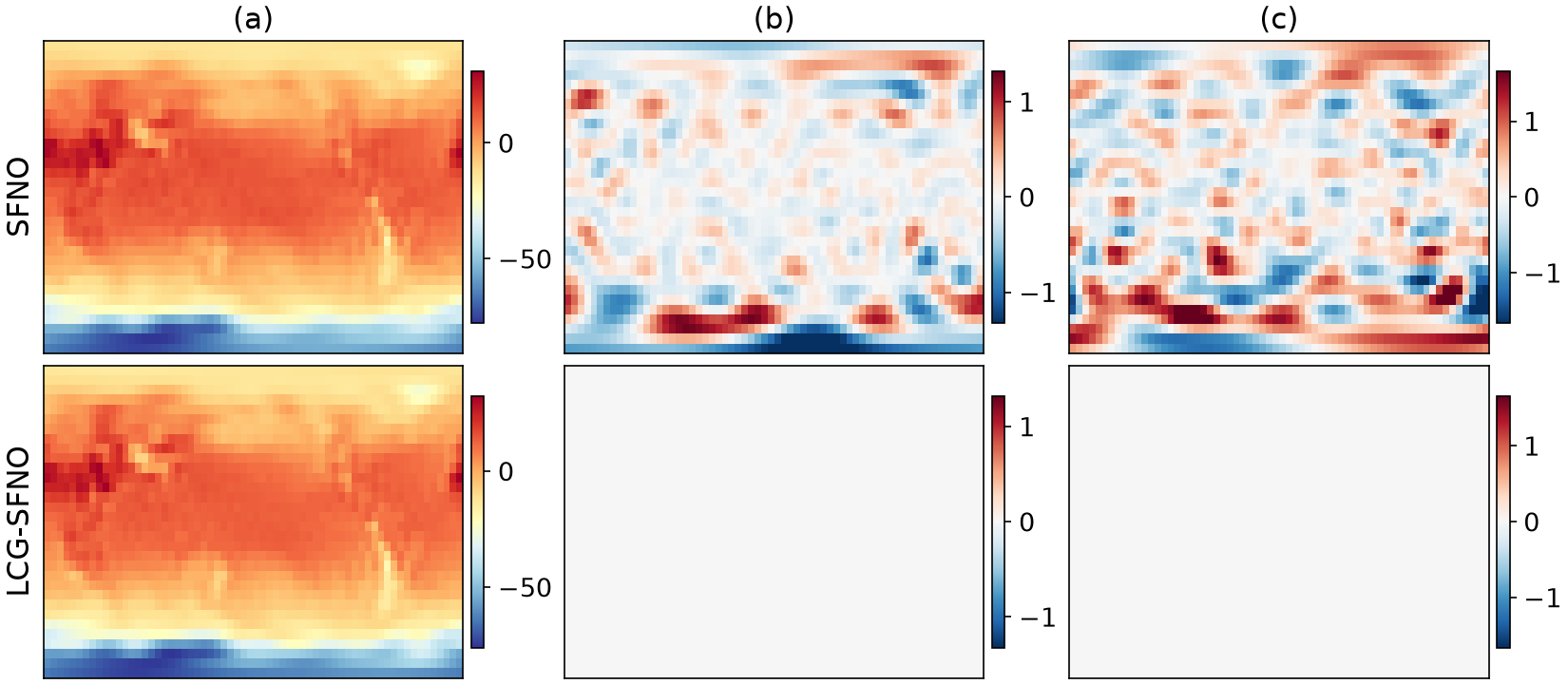}
\vspace{-3mm}
\caption{\textbf{Spatial structure of rotational equivariance error.}
WeatherBench results for one test state $\mathbf{x}$ and one random rotation
$R$. (a) One-day 2\,m temperature forecast; (b,c) equivariance residual
$R^{-1}\mathcal{G}(R\mathbf{x},R\mathbf{c})
-\mathcal{G}(\mathbf{x},\mathbf{c})$
for 2\,m temperature and 10\,m zonal wind. Rows correspond to SFNO and
LCG-SFNO, with shared color scales within each column. The structured residual
visible for SFNO is absent at the displayed scale for LCG-SFNO.}
\label{fig:equivmaps}
\end{figure}


In this section, we evaluate LCG-SFNO on the spherical shallow-water equations
and WeatherBench. We assess rotational equivariance, the contributions of the
main architectural components, computational scalability, and forecast
performance. Additional implementation details and ablations are provided in
the appendix.

\subsection{Benchmarks and Evaluation Protocol}
\label{sec:setup}

We evaluate on two spherical forecasting benchmarks: the spherical
shallow-water equations (SSWE) and WeatherBench. SSWE provides a controlled
setting for measuring rotational equivariance and autoregressive rollout error.
WeatherBench at $5.625^\circ$ resolution, constructed from ERA5 reanalysis,
provides a global atmospheric forecasting task. We train separate direct
WeatherBench models at lead times of one, three, and five days and report
anomaly correlation coefficient (ACC) and normalized latitude-weighted
mean-square error (MSE). On SSWE, we report relative $L^2$ rollout error.

We measure rotational equivariance using the relative equivariance error
\begin{equation}
    \epsilon_{\mathrm{eq}}
    =
    \frac{
    \left\|
    \mathcal{G}(R\mathbf{x},R\mathbf{c})
    -
    R\mathcal{G}(\mathbf{x},\mathbf{c})
    \right\|_2
    }{
    \left\|
    R\mathcal{G}(\mathbf{x},\mathbf{c})
    \right\|_2
    },
    \label{eq:equivariance-metric}
\end{equation}
where $\mathcal{G}$ denotes the forecasting operator and $R\in\SO(3)$ is
applied jointly to the input state and auxiliary spatial fields. Rotations are
performed directly in spherical-harmonic coefficient space, avoiding spatial
interpolation. We average the error over eight random rotations and sixteen test
states; further details are given in Appendix~\ref{app:protocol}.

We compare with ClimaX~\citep{nguyen2023climax},
FourCastNet~\citep{pathak2022fourcastnet}, SHNet with the GSNO
operator~\citep{tang2026gsno}, and SFNO~\citep{bonev2023sfno}. All models are
trained using the same data splits and evaluation grids. Dataset details,
optimization settings, and model configurations are given in
Appendix~\ref{app:protocol}.

\subsection{Rotational Equivariance}
\label{sec:equivariance-results}

\begin{table}[t]
\caption{\textbf{Incremental component ablation on one-day WeatherBench.}
Rows progressively add the full-band backbone, CG coupling, and oversampled
nonlinear pathway. ``Epochs'' denotes the number of epochs required to reach
row B's best validation error. Rows A--C report mean $\pm$ standard deviation
over three seeds; row D is a single-seed diagnostic.}
\label{tab:component-ablation}
\centering
\vspace{2mm}
\small
\begin{tabular}{lcccc}
\toprule
Model & ACC & MSE & Epochs & Equivariance \\
\midrule
A~~SFNO reference
    & $84.0\pm0.0$ & $187.7\pm0.2$ & never & $9.5\times10^{-2}$ \\
B~~+ full-band backbone
    & $93.4\pm0.1$ & $81.5\pm0.6$ & 39 & $6.7\times10^{-2}$ \\
C~~+ CG coupling
    & $93.7\pm0.0$ & $77.9\pm0.1$ & 13 & $6.8\times10^{-2}$ \\
D~~+ oversampled nonlinear pathway
    & $93.4$ & $80.7$ & 21 & $7.2\times10^{-3}$ \\
\bottomrule
\end{tabular}
\vspace{-3mm}
\end{table}

\begin{figure}[b]
\vspace{-6mm}
\centering
\includegraphics[width=\linewidth]{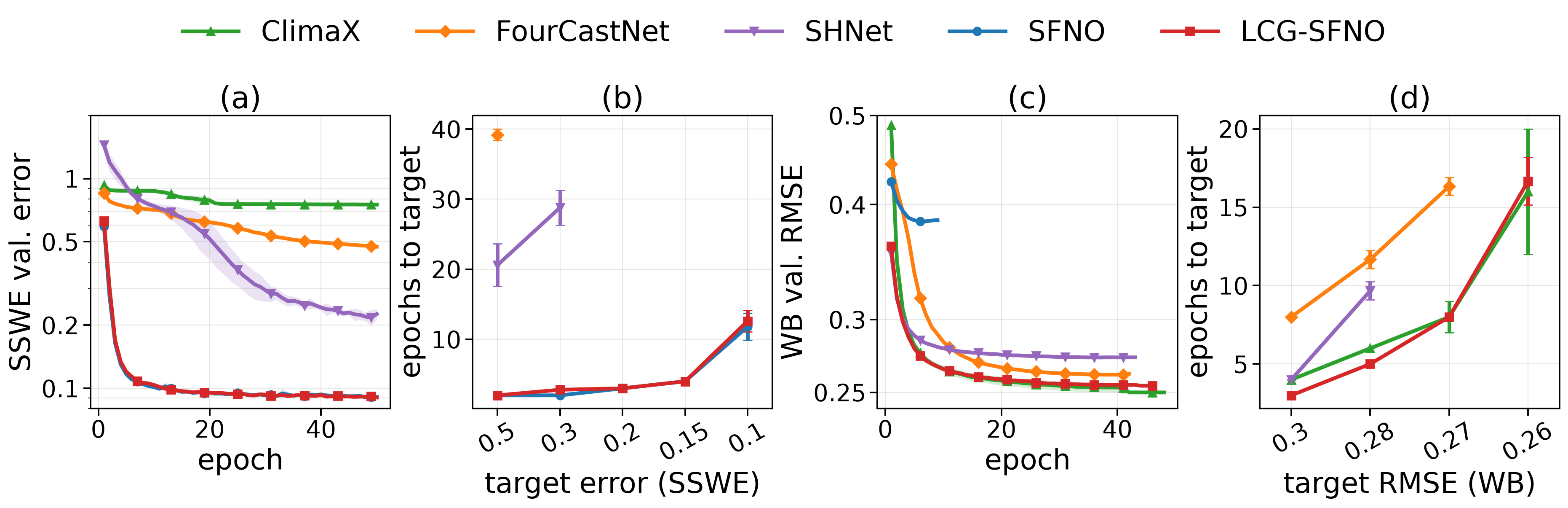}
\vspace{-8mm}
\caption{\textbf{Training efficiency.} \textbf{(a,b)} SSWE one-step validation error and epochs to reach each error target over five seeds. \textbf{(c,d)} Corresponding results for one-day WeatherBench over three seeds. Curves terminate when a model fails to reach a stricter target or early stopping occurs. LCG-SFNO matches SFNO's convergence on SSWE, reaches the intermediate WeatherBench targets in fewer epochs, and matches the best model at the strictest target.}
\label{fig:curves}
\end{figure}

Table~\ref{tab:equiv} shows that LCG-SFNO reduces end-to-end equivariance error
from \(9.5\times10^{-4}\) to \(8.2\times10^{-7}\) on SSWE and from approximately
\(10^{-1}\) to \(10^{-6}\) on WeatherBench, reductions of more than three and
approximately five orders of magnitude, respectively. The resulting errors are
at numerical precision. This contrast with SFNO shows that a
spherical-harmonic representation alone is insufficient to guarantee
end-to-end rotational equivariance. Figure~\ref{fig:equivmaps} further shows
that SFNO exhibits spatially structured residuals near fronts and
high-latitude gradients, whereas the LCG-SFNO residual is not visible on the
same color scale.

\subsection{Component Effects and Optimization}
\label{sec:components}

Table~\ref{tab:component-ablation} isolates the contributions of the main
architectural changes on one-day WeatherBench. The A$\rightarrow$B transition
accounts for most of the forecast improvement over the reference SFNO. Adding
CG coupling in B$\rightarrow$C further improves forecast error and reduces the
number of epochs required to reach row B's best validation error from $39$ to
$13$, while leaving equivariance error essentially unchanged. The single-seed
C$\rightarrow$D diagnostic instead reduces equivariance error from
$6.8\times10^{-2}$ to $7.2\times10^{-3}$, with only a modest change in forecast
accuracy. The ablation therefore separates the roles of the three components:
the full-band backbone provides most of the accuracy gain, CG coupling improves
forecasting and optimization, and the oversampled nonlinear pathway reduces
discretization-induced equivariance error.

Figure~\ref{fig:curves} shows the corresponding optimization behavior over the
full training trajectories. SFNO and LCG-SFNO converge similarly on SSWE,
whereas LCG-SFNO reaches intermediate WeatherBench error targets in fewer epochs
than the comparison models. Together with the B$\rightarrow$C ablation, this
supports attributing part of the faster WeatherBench convergence to the CG
interaction rather than to the stronger spectral backbone alone.

The final model replaces the non-polynomial activation used in this diagnostic
with the quadratic activation and exact spectral projection described in
Section~\ref{sec:dealiasing}.



\subsection{Computational Scalability}
\label{sec:speed}

\begin{figure}[b]
\vspace{-5mm}
\centering
\includegraphics[width=0.75\linewidth]{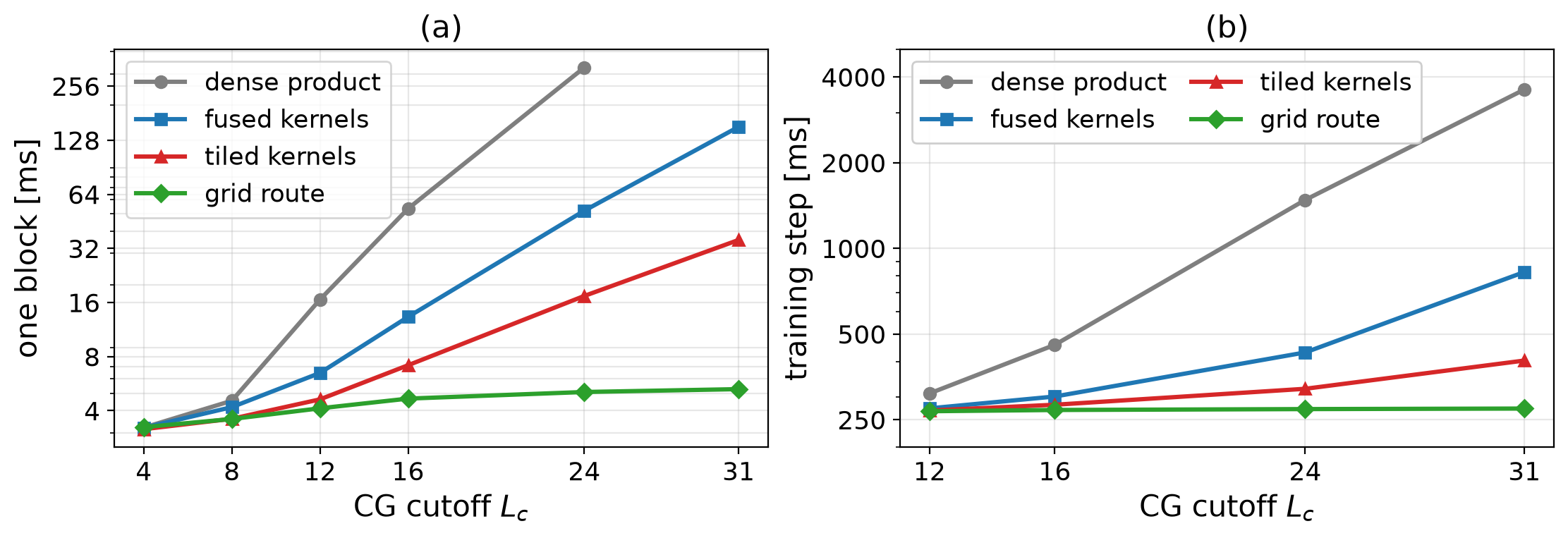}
\vspace{-4mm}
\caption{\textbf{Computational cost versus coupled bandwidth.} \textbf{(a)} CG-block forward-and-backward time; \textbf{(b)} end-to-end training-step time, both versus input cutoff \(L_c\). The grid route remains nearly constant with increasing bandwidth.}
\label{fig:speed}
\end{figure}

\begin{table}[t]
  \caption{\textbf{CG-block cost versus coupled bandwidth.} Forward-and-backward time (ms) for one WeatherBench CG block (186 channels, rank $r=4$, batch 64, single precision). LCG-SFNO uses the \(12/24\) band. Direct implementations become more expensive with bandwidth, whereas the grid route remains nearly constant. \(^{\dagger}\)The \(31/31\) dense product exceeds 96\,GB at batch 64 and is timed using four micro-batches of 16.}
  \label{tab:cgblock}
  \vspace{2mm}
  \centering
  \small
  \setlength{\tabcolsep}{5pt}
  \begin{tabular}{lrcccc}
    \toprule
    Band $L_c/L_o$ & Triples & Dense & Fused & Tiled & Grid route \\
    \midrule
    4/8   & 85      & 3.2 & 3.2 & 3.2 & 3.2 \\
    8/16  & 489     & 4.5 & 4.2 & 3.6 & 3.6 \\
    12/24 & 1{,}469 & 16.6 & 6.5 & 4.6 & 4.1 \\
    16/31 & 3{,}280 & 53.2 & 13.3 & 7.2 & 4.7 \\
    24/31 & 9{,}456 & 325 & 51.9 & 17.4 & 5.1 \\
    31/31 & 16{,}400 & 855$^\dagger$ & 152 & 35.6 & 5.3 \\
    \bottomrule
  \end{tabular}
  \vspace{-5mm}
\end{table}

Table~\ref{tab:cgblock} and Figure~\ref{fig:speed} show two computational
regimes. At the \(12/24\) operating band, the tiled direct implementation costs
\(4.6\) ms per block while retaining an independent learned weight for every
admissible degree path. As the coupled bandwidth increases, its cost rises to
\(35.6\) ms at \(31/31\), whereas the factorized grid route remains nearly
constant. The tiled direct route is therefore effective at moderate bandwidths,
while the factorized route provides better scaling for wider coupling.

In a single-seed WeatherBench comparison, widening the coupled band from \(12/24\) to \(31/31\) provides little additional
forecast benefit: one-day ACC changes from \(93.4\) to \(93.5\) and MSE from
\(80.7\) to \(80.5\), while tiled CG cost increases by nearly eightfold. We
therefore use the \(12/24\) band in the main model. Appendix
Figure~\ref{fig:spectrum} provides the corresponding spectral analysis of the
WeatherBench fields.

\subsection{Forecast Performance}
\label{sec:accuracy}

On SSWE, LCG-SFNO matches SFNO within variation across seeds at one, five, and
ten hours while reducing equivariance error by more than three orders of
magnitude. Thus, enforcing substantially stronger rotational consistency does
not degrade forecast accuracy in this controlled setting.

On WeatherBench, LCG-SFNO substantially improves over SFNO at a matched
parameter count. At one day, ACC increases from \(84.0\) to \(93.3\), while
normalized MSE decreases from \(187.7\) to \(82.1\). LCG-SFNO attains the
highest mean one-day ACC among the evaluated models, while ClimaX performs
better at three days and SHNet at five days. Tables~\ref{tab:sswe}
and~\ref{tab:wb} report the complete results. Variable-wise scores, rollout
trajectories, and additional architecture and transfer ablations are provided
in Appendix~\ref{app:pervar}--\ref{app:transfer}.

\begin{table}[t]
  \caption{\textbf{Shallow-water forecasting.} Relative \(L^2\) rollout error (\(\times10^{-3}\)) over 50 test initial conditions, reported as mean \(\pm\) standard deviation over five seeds. LCG-SFNO matches SFNO within seed variation across all horizons while using the fewest parameters.}
  \label{tab:sswe}
  \vspace{2mm}
  \centering
  \small
  \begin{tabular}{lrccc}
    \toprule
    Model & Params & 1 h & 5 h & 10 h \\
    \midrule
    ClimaX                   & 1.96M & 421 $\pm$ 0 & 474 $\pm$ 5 & 535 $\pm$ 10 \\
    FourCastNet              & 1.66M & 285 $\pm$ 2 & 687 $\pm$ 40 & 1309 $\pm$ 110 \\
    SHNet (GSNO)             & 0.87M & 142 $\pm$ 6 & 391 $\pm$ 16 & 625 $\pm$ 189 \\
    SFNO                     & \second{0.28M} & \second{58.0 $\pm$ 1.2} & \second{121 $\pm$ 6} & \second{219 $\pm$ 22} \\
    \textbf{LCG-SFNO (ours)} & \textbf{0.27M} & \textbf{57.4 $\pm$ 1.4} & \textbf{119 $\pm$ 7} & \textbf{211 $\pm$ 17} \\
    \bottomrule
  \end{tabular}
  \vspace{-3mm}
\end{table}

\begin{table}[b]
  \caption{\textbf{WeatherBench forecasting.} ACC (\%) and normalized latitude-weighted MSE (\(\times10^{-3}\)), reported as mean \(\pm\) standard deviation over three seeds. LCG-SFNO substantially improves over SFNO across all horizons and achieves the highest mean one-day ACC; ClimaX and SHNet lead on the remaining aggregate metrics.}
  \label{tab:wb}
  \centering
  \vspace{2mm}
  \small
  \setlength{\tabcolsep}{3.5pt}
  \resizebox{\textwidth}{!}{%
  \begin{tabular}{lrcccccc}
    \toprule
    & & \multicolumn{2}{c}{1 day} & \multicolumn{2}{c}{3 days} & \multicolumn{2}{c}{5 days} \\
    \cmidrule(lr){3-4}\cmidrule(lr){5-6}\cmidrule(lr){7-8}
    Model & Params & ACC & MSE & ACC & MSE & ACC & MSE \\
    \midrule
    ClimaX                   & 5.40M & \second{93.2 $\pm$ 0.3} & \textbf{76.8 $\pm$ 0.5} & \textbf{78.3 $\pm$ 0.4} & \textbf{242.4 $\pm$ 2.5} & \second{53.1 $\pm$ 0.8} & 435.0 $\pm$ 4.5 \\
    FourCastNet              & 5.60M & 92.8 $\pm$ 0.1 & 82.8 $\pm$ 0.3 & 69.3 $\pm$ 0.1 & 325.1 $\pm$ 0.9 & 50.5 $\pm$ 0.1 & 440.1 $\pm$ 0.1 \\
    SHNet (GSNO)             & 4.03M & 92.4 $\pm$ 0.0 & 92.7 $\pm$ 0.1 & \second{74.2 $\pm$ 0.1} & \second{281.6 $\pm$ 0.9} & \textbf{54.9 $\pm$ 0.1} & \textbf{420.1 $\pm$ 1.1} \\
    SFNO                     & 5.26M & 84.0 $\pm$ 0.0 & 187.7 $\pm$ 0.2 & 65.4 $\pm$ 0.2 & 350.8 $\pm$ 1.4 & 48.6 $\pm$ 0.1 & 451.1 $\pm$ 0.6 \\
    \textbf{LCG-SFNO (ours)} & 5.24M & \textbf{93.3 $\pm$ 0.1} & \second{82.1 $\pm$ 1.4} & 73.5 $\pm$ 0.1 & 286.5 $\pm$ 0.9 & 51.7 $\pm$ 0.2 & \second{430.4 $\pm$ 1.4} \\
    \bottomrule
  \end{tabular}}
\end{table}

\section{Conclusion}

A spherical-harmonic representation alone does not guarantee end-to-end
rotational equivariance, while full-band Clebsch--Gordan (CG) coupling is
computationally expensive. LCG-SFNO combines full-band degree-wise spectral
processing with low-rank CG interactions over a controlled low-degree band and
a quadratic nonlinearity with exact spectral projection. SSHCG provides
efficient direct and factorized implementations of the CG interaction.

LCG-SFNO reduces equivariance error to numerical precision on both spherical
shallow-water dynamics and WeatherBench, while matching SFNO on the former and
substantially improving forecast accuracy on the latter. Ablations show distinct
roles for the spectral backbone, CG coupling, and dealiased nonlinear pathway,
while bandwidth experiments show little benefit from extending CG coupling
across the full spectrum. These results demonstrate end-to-end rotationally
equivariant spherical forecasting without full-band CG interactions.

\clearpage
\subsection*{Reproducibility Statement}

All models in this paper, including the baselines, were trained by us under one
protocol, and every element needed to repeat the experiments is described in the
paper. Appendix~\ref{app:protocol} specifies the two benchmarks, the data splits
and grids, the optimizer, schedule, batch size, early-stopping rule, and the
number of seeds behind every reported mean and standard deviation, and
Appendix~\ref{app:hyper} lists the architecture of each compared model at the
sizes of the benchmark paper. Appendix~\ref{app:background} derives the spherical
representation and the dealiasing rule that makes the quadratic activation
exact, and Appendix~\ref{app:method} gives the low-rank Clebsch--Gordan
parameterization, the spectral normalization, the encoder and decoder, and the
equivariance of each learned operation. The rotational-consistency measurements
follow the diagnostic of Appendix~\ref{app:equivariance}, which states the
rotation sampling, the band-limiting convention, and the numerical precision
used. Appendix~\ref{app:acceleration} describes the kernels and the factorized
grid route, their complexity, and their numerical agreement with the dense
reference. 

\subsection*{Ethics Statement}

This work studies operator learning for physical forecasting and uses only
public data: the ERA5 reanalysis through the WeatherBench benchmark and
trajectories generated by a public shallow-water solver. No human subjects,
personal data, or sensitive attributes are involved. Data-driven weather models
are inexpensive to evaluate once trained, which can broaden access to forecast
guidance; they are also statistical models that can fail outside their training
distribution, and the rotational consistency studied here addresses only one of
the properties a forecast system should have before operational use. We see no
other ethical concerns specific to this work.

\subsection*{AI Use Statement}

In this work, we used generative AI tools for generating synthetic datasets, implementing methods, supporting qualitative and thematic data analysis, and designing and running experiments. We have not used generative AI tools for developing theoretical models or conceptual frameworks, formulating mathematical claims, providing critical ingredients for proving mathematical claims, or interpreting results, and assisting in writing proofs, proposing or refining hypotheses, assisting with translation, cleaning and reformatting datasets are not applicable to this work. Additionally, we used generative AI tools for creating or editing software code, drafting parts of the manuscript, and editing the manuscript to improve readability. We have reviewed all AI-assisted work. The authors reviewed the implementation and experimental procedures, verified the reported results, and checked the manuscript for accuracy. We take responsibility for the final content of this work, including text, claims or artifacts produced with the aid of generative AI.

\bibliographystyle{iclr2027_conference}
\bibliography{iclr2027_conference}

\clearpage
\appendix



\begingroup
\def\AppendixBackground{1}

\ifdefined\AppendixBackground
\section{Spherical Harmonics and Clebsch--Gordan Products in Detail}
\label{app:background}

This appendix derives the facts that Section~\ref{sec:background} states and
that the exactness claims of the method rest on: the quadrature on which the
transforms are exact, the action of rotations on coefficients, the
Clebsch--Gordan series behind the product of two harmonics, the equivariance of
the bilinear coupling and its cost, and the dealiasing rule for a polynomial
activation. Standard references are \citet{edmonds1996angular} for the algebra
and \citet{driscoll1994computing,boyd2001chebyshev} for the numerics.

\subsection{Harmonics, band limits, and quadrature}
\label{app:harmonics}

The spherical harmonics are the eigenfunctions of the Laplace--Beltrami
operator on the unit sphere, $\Delta_{S^2}Y_{\ell m}=-\ell(\ell+1)Y_{\ell m}$.
In the complex form with the Condon--Shortley phase they read
\begin{equation}
  Y_{\ell}^{m}(\theta,\varphi)
  =\sqrt{\frac{2\ell+1}{4\pi}\frac{(\ell-m)!}{(\ell+m)!}}\;
  P_{\ell}^{m}(\cos\theta)\,e^{im\varphi},
  \qquad 0\le|m|\le\ell,
  \label{eq:sh-definition}
\end{equation}
with the associated Legendre functions $P_\ell^m$, and they are orthonormal,
$\int_{S^2}Y_\ell^m\,\overline{Y_{\ell'}^{m'}}\,\mathrm{d}\Omega
=\delta_{\ell\ell'}\delta_{mm'}$. The eigenspace of degree $\ell$ has
dimension $2\ell+1$, and the harmonics of degree at most $L$ span the
$(L+1)^2$-dimensional space of band-limited functions. For a real field the
coefficients of orders $m$ and $-m$ are complex conjugates up to sign, so the
real basis $\{Y_{\ell 0},\sqrt2\,\mathrm{Re}\,Y_\ell^m,\sqrt2\,\mathrm{Im}\,Y_\ell^m\}_{m>0}$
carries the same information with real coefficients; the model stores the
half-spectrum $m\ge0$ of this real basis, and every statement below holds in
either basis after the unitary change of basis between them.

The transforms are computed on a Legendre--Gauss grid: $n_\theta$ colatitudes
at the roots of the Legendre polynomial $P_{n_\theta}(\cos\theta)$ with
Gauss--Legendre weights, and $n_\varphi$ equispaced longitudes. Two exactness
properties of this grid carry the whole construction. In longitude, the
trapezoidal rule with $n_\varphi$ points integrates $e^{ik\varphi}$ exactly
unless $k$ is a nonzero multiple of $n_\varphi$; in colatitude, the
Gauss--Legendre rule with $n_\theta$ nodes integrates every polynomial of
degree at most $2n_\theta-1$ in $\cos\theta$ exactly. The product of two
harmonics of degrees $a$ and $b$ is, in colatitude, a polynomial of degree at
most $a+b$ in $\cos\theta$ (the half-integer powers of $\sin\theta$ that
$P_\ell^m$ carries pair up because the orders add). Hence the analysis
$u_{\ell m}=\sum_{ij}w_{ij}\,u(\theta_i,\varphi_j)\,\overline{Y_\ell^m(\theta_i,\varphi_j)}$
of a field band-limited at $L$ is exact whenever $2L\le 2n_\theta-1$ and
$2L<n_\varphi$, that is, for $n_\theta\ge L+1$ and $n_\varphi\ge 2L+1$. The
$32\times64$ WeatherBench grid therefore represents degrees up to $31$
exactly, and the analysis and synthesis $\cS$, $\cS^{-1}$ of the main text
are inverse to each other on that band.

\subsection{Rotations and Wigner $D$-matrices}
\label{app:wigner}

A rotation $R\in\SO(3)$ acts on functions by $(Ru)(x)=u(R^{-1}x)$. Because
$\Delta_{S^2}$ commutes with rotations, each eigenspace is mapped to itself,
so the action never mixes degrees; on the degree-$\ell$ eigenspace it is
represented by the unitary $(2\ell+1)\times(2\ell+1)$ Wigner $D$-matrix
$D^\ell(R)$, which in $ZYZ$ Euler angles $(\alpha,\beta,\gamma)$ factorizes as
$D^\ell_{mm'}(\alpha,\beta,\gamma)=e^{-im\alpha}\,d^\ell_{mm'}(\beta)\,e^{-im'\gamma}$
with the real Wigner $d$-functions; see the textbooks of \citet{edmonds1996angular}
and \citet{varshalovich1988quantum} for the construction and properties of
$D^\ell(R)$. These
representations are irreducible: no proper subspace of a degree is invariant
under all rotations, which is the statement that the degrees are the smallest
building blocks that rotations respect. Equation~\eqref{eq:wigner} of the main
text is this action written in coefficients. Our equivariance diagnostic
(Appendix~\ref{app:equivariance}) builds $D^\ell(R)$ as the matrix exponential
of the angular-momentum generators in the real basis and applies it degree by
degree, so a field is rotated exactly on its band without interpolation on the
grid.

Two consequences follow directly. A map that acts on the coefficients of each
degree by the same matrix for every order, $u_{\ell m}\mapsto\sum_b
W_{\ell ab}\,u_{b\ell m}$, commutes with $D^\ell(R)$ because $D^\ell(R)$ acts on
the order index and $W_\ell$ on the channel index; this is the equivariance of
the per-degree linear map \eqref{eq:degree-linear}. Conversely, a weight that
depends on $m$, or a complex weight acting on the half-spectrum (which acts as
$w$ on $m>0$ and $\bar w$ on $m<0$ of the real field), does not commute with
$D^\ell(R)$ in general, which is why the model uses real weights shared across
orders.

\subsection{The Clebsch--Gordan series and the product of two harmonics}
\label{app:cg-series}

The tensor product of the irreducible representations of degrees $\ell_1$ and
$\ell_2$ is reducible, and it decomposes into exactly one copy of each degree
between $|\ell_1-\ell_2|$ and $\ell_1+\ell_2$,
\begin{equation}
  D^{\ell_1}(R)\otimes D^{\ell_2}(R)
  \;\cong\;
  \bigoplus_{\ell=|\ell_1-\ell_2|}^{\ell_1+\ell_2} D^{\ell}(R).
  \label{eq:cg-series}
\end{equation}
The Clebsch--Gordan coefficients $C^{\ell m}_{\ell_1m_1,\ell_2m_2}$ are the
entries of the unitary matrix that realizes this decomposition: the vectors
$\sum_{m_1,m_2}C^{\ell m}_{\ell_1m_1,\ell_2m_2}\,e_{\ell_1m_1}\otimes e_{\ell_2m_2}$
form a basis of the degree-$\ell$ summand, and they vanish unless
$m=m_1+m_2$ and \eqref{eq:cg-selection} holds. This is the source of the
selection rules in Section~\ref{sec:background}: the triangle inequality is
the range of \eqref{eq:cg-series}, and $m=m_1+m_2$ is the addition of the
$z$-components.

The product of two harmonics is again a finite sum of harmonics because the
product of functions on the two eigenspaces lives in their tensor product,
which \eqref{eq:cg-series} decomposes:
\begin{equation}
  Y_{\ell_1}^{m_1}\,Y_{\ell_2}^{m_2}
  =\sum_{\ell=|\ell_1-\ell_2|}^{\ell_1+\ell_2}
  \sqrt{\frac{(2\ell_1+1)(2\ell_2+1)}{4\pi(2\ell+1)}}\;
  C^{\ell 0}_{\ell_10,\ell_20}\;
  C^{\ell\,m_1+m_2}_{\ell_1m_1,\ell_2m_2}\;
  Y_{\ell}^{m_1+m_2}.
  \label{eq:gaunt-expansion}
\end{equation}
The coefficients on the right are the Gaunt coefficients, the integrals of
three harmonics over the sphere. The factor $C^{\ell0}_{\ell_10,\ell_20}$
vanishes unless $\ell_1+\ell_2+\ell$ is even, so the product of two fields
couples only degree triples of even parity, while the learned coupling of
\eqref{eq:cg-coupling} keeps the odd triples as well and lets the path
weights $\gamma$ decide their use. Equation~\eqref{eq:gaunt-expansion} also
proves the band-limit statement used throughout the paper: the product of
fields band-limited at $L_1$ and $L_2$ is band-limited at $L_1+L_2$, and the
square of a degree-$L$ field at $2L$.

\subsection{Equivariance and cost of the Clebsch--Gordan product}
\label{app:cg-equivariance}

The bilinear map \eqref{eq:cg-product} is equivariant because the CG
coefficients intertwine the representations. Writing $p'=D^{\ell_1}(R)p$ and
$q'=D^{\ell_2}(R)q$ for the rotated inputs, the defining property of the
unitary change of basis in \eqref{eq:cg-series} gives
\begin{equation}
  \sum_{m_1,m_2}C^{\ell m}_{\ell_1m_1,\ell_2m_2}\,p'_{\ell_1m_1}\,q'_{\ell_2m_2}
  =\sum_{m'}D^{\ell}_{mm'}(R)
  \sum_{m_1,m_2}C^{\ell m'}_{\ell_1m_1,\ell_2m_2}\,p_{\ell_1m_1}\,q_{\ell_2m_2},
  \label{eq:cg-intertwine}
\end{equation}
that is, $z(D^{\ell_1}p,D^{\ell_2}q)=D^{\ell}z(p,q)$ for every degree $\ell$ of
the output. The identity holds for every rotation and involves no grid, so
the coupling is exactly equivariant however coarse the data grid is; the
learned parts of \eqref{eq:cg-coupling}, the projections $P$, $Q$, $O$ and the
path weights $\gamma$, act on the channel and degree-triple indices, which
$D^\ell(R)$ does not touch, and therefore preserve the identity.

The cost follows from the same structure. For a coupled band with input
cutoff $L_c$ and output cutoff $L_o\ge 2L_c$, the number of admissible triples is
\begin{equation}
  N_{\mathrm{triples}}
  =\sum_{\ell_1=0}^{L_c}\sum_{\ell_2=0}^{L_c}
  \bigl(\min(\ell_1+\ell_2,L_o)-|\ell_1-\ell_2|+1\bigr)
  =\sum_{\ell_1,\ell_2\le L_c}\bigl(2\min(\ell_1,\ell_2)+1\bigr)
  \sim\tfrac{2}{3}L_c^{3},
  \label{eq:triple-count}
\end{equation}
which gives the 1{,}469 triples of the $12/24$ band and the 16{,}400 of the
full band in Table~\ref{tab:cgblock}. Within one triple the constraint
$m=m_1+m_2$ leaves about $(2\ell_1+1)(2\ell_2+1)$ nonzero couplings, so the
dense contraction touches $O(L_c^{5})$ angular coefficients per channel and
batch element, and a dense implementation that stores every $(m,m_1,m_2)$
combination stores far more zeros than nonzeros. This growth is what the
kernels of Appendix~\ref{app:sshcg-direct} organize around, and
\eqref{eq:gaunt-expansion} is what the grid route of
Appendix~\ref{app:gaunt} exploits: for path weights that factorize over the
degrees, the contraction equals a synthesis of the two inputs, a pointwise
product on a grid that resolves degree $2L_c$, and an analysis, all three of
which are exact by Appendix~\ref{app:harmonics}.

\subsection{The dealiasing rule for a polynomial activation}
\label{app:dealiasing-rule}

Let $u$ be band-limited at $L$ and let $\sigma$ be a polynomial of degree $p$.
By \eqref{eq:gaunt-expansion}, $\sigma(u)$ is band-limited at $pL$, so it is a
finite expansion and no truncation is involved in principle; the only
question is whether the projection $\Pi_L\cS_f[\sigma(u)]$ onto degrees
$\ell\le L$ in \eqref{eq:grid-update} is computed exactly by the quadrature
of the oversampled grid. Its integrand is a product of a degree-$pL$ and a
degree-$L$ harmonic, hence of degree $(p+1)L$ in $\cos\theta$, and its
longitudinal orders satisfy $|m_1+\dots+m_p-m|\le(p+1)L$. Exactness therefore
requires
\begin{equation}
  2n_\theta'-1\ \ge\ (p+1)L
  \qquad\text{and}\qquad
  n_\varphi'\ >\ (p+1)L,
  \label{eq:dealias-rule}
\end{equation}
that is, an oversampling factor $f\ge(p+1)/2$ relative to the $n_\theta=L+1$,
$n_\varphi=2L+2$ grid that represents the band itself. For the quadratic
activation this is the three-halves rule of \citet{orszag1971elimination},
$n_\theta'\ge(3L+1)/2$ and $n_\varphi'>3L$: for $L=31$ the $48\times96$ grid of
the blocks satisfies both with $n_\theta'=48\ge47$ and $n_\varphi'=96>93$. The
content of $\sigma(u)$ above degree $L$ is not represented on that grid, but
it is not needed: the rule guarantees that it does not alias into the retained
band, which is all the projection requires. For a non-polynomial activation
the expansion of $\sigma(u)$ is infinite, the integrand is not a polynomial,
and no finite grid makes the projection exact; the aliased remainder decreases
with $f$ but never vanishes, which is the residual measured in
Appendix~\ref{app:oversampling}. The decoder retains the fine band rather than
projecting to $L$, so its grid must represent the product itself; for the
quadratic this is degree $2L$, which the $2\times$ codec grid with
$n_\theta'=64$ and $n_\varphi'=128$ resolves.

\fi

\ifdefined\AppendixArchitecture

\section{Architecture and Equivariance Details}
\label{app:method}

This appendix provides the implementation details underlying the equivariant
construction in Section~\ref{sec:method}. We first specify the spherical-harmonic
representation and the equivariant degree-wise linear maps, then describe the
low-rank Clebsch--Gordan parameterization, the treatment of aliasing and spectral
normalization, and finally the encoder, decoder, and output representation.

\subsection{Spherical representation and equivariant degree-wise maps}
\label{app:spherical}

A real spherical field $u$ is represented by spherical-harmonic coefficients
$u_{\ell m}$ obtained with the Legendre--Gauss transform $\cS$ and reconstructed
with $\cS^{-1}$. Under a rotation $R\in\SO(3)$, the coefficients of fixed degree
$\ell$ transform as
\begin{equation}
  u_\ell \mapsto D^\ell(R)u_\ell .
\end{equation}
Consequently, any channel transformation that is shared across the orders $m$
within a fixed degree, including Eq.~\eqref{eq:degree-linear}, commutes with
$D^\ell(R)$ and is therefore equivariant.

We use real-valued per-degree weights throughout. For a real field represented
through its non-negative-$m$ half-spectrum, an unrestricted complex weight acts
as $w$ on $m>0$ and $\bar w$ on $m<0$. This does not, in general, define the same
channel transformation across the full degree-$\ell$ irreducible representation
and therefore need not commute with rotations. Restricting the per-degree weight
to be real applies the same channel transformation across all orders and
preserves equivariance.

\subsection{Low-rank Clebsch--Gordan parameterization}
\label{app:cg-param}

The channel dependence of the CG correction is factorized through rank $r$.
Two learned projections first map the hidden channels into rank-indexed streams,
\begin{align}
  p^{(s)}_{\ell m} &= \sum_b P_{\ell sb}H_{b\ell m}, &
  q^{(s)}_{\ell m} &= \sum_b Q_{\ell sb}H_{b\ell m},
  \label{eq:low-rank-streams}
\end{align}
and Eq.~\eqref{eq:cg-coupling} couples each stream over all admissible degree
triples, in the manner of the Clebsch--Gordan and tensor-product layers of
\citet{kondor2018clebsch}, \citet{thomas2018tensor}, and
\citet{weiler2018steerable}. The result is projected back to model channels as
\begin{equation}
  \Delta H^{\mathrm{CG}}_{a\ell m}
  = \alpha_\ell\sum_s O_{\ell as}z^{(s)}_{\ell m},
  \label{eq:cg-output}
\end{equation}
where $\alpha_\ell$ is initialized at zero. Inputs to the tensor product are
restricted to $\ell_1,\ell_2\le L_c$, and the correction is written only to
$\ell\le L_o$; degrees above $L_o$ therefore follow the original SFNO pathway.
Because the learned channel projections act independently within each
irreducible representation and the CG tensor product is an equivariant bilinear
map between irreducible representations~\citep{edmonds1996angular,varshalovich1988quantum},
the complete correction is exactly $\SO(3)$-equivariant. The rank-$r$
factorization plays the role of the learned channel-mixing weights of a general
tensor-product layer~\citep{geiger2022e3nn}, with the channel dimension of the
product reduced from the model width to $r$; other routes to a cheaper product
restrict the degree paths or change the evaluation basis
\citep{passaro2023reducing,luo2024gaunt}, whereas this factorization keeps every
admissible path and evaluates it directly.

\subsection{Dealiased nonlinear evaluation and spectral normalization}
\label{app:dealiasing}

Pointwise nonlinearities are equivariant as operations on continuous spherical
fields, but their discrete evaluation can introduce aliasing when the generated
harmonics exceed the retained spectral band. For a polynomial of degree $p$
applied to a degree-$L$ field, the output contains harmonics up to degree $pL$,
leading to the classical oversampling condition
$f\ge (p+1)/2$~\citep{orszag1971elimination}. The main model uses the quadratic
$\sigma(x)=x/2+x^{2}/4$, for which $f=1.5$ makes the projection exact. A
non-polynomial activation such as GELU generates an infinite series, no finite
oversampling factor removes all of it, and the remaining aliasing can only be
controlled empirically through $f$. Fused into a single kernel, the quadratic
costs the same as GELU at equal grid size (a WeatherBench training step of
176\,ms for either at batch 64 under mixed precision, with the blocks at
$f=1.5$ and the codec at $f=2$; a shallow-water step of 127\,ms at $f=3$ for
either), and
at its exact grid $f=1.5$ the shallow-water step drops to 44\,ms with a third
of the memory.

Using the equivariance diagnostic of Appendix~\ref{app:equivariance}, a single
randomly initialized block has relative equivariance error $10^{-6}$ for the
quadratic activation at $f=2$, $2.5\times10^{-4}$ for GELU at $f=2$, and
$6\times10^{-6}$ for GELU at $f=3$, compared with $10^{-1}$ for GELU evaluated
on the data grid. For the complete WeatherBench network at initialization, the
quadratic gives $6.8\times10^{-7}$ at the grids of the main model (blocks
$f=1.5$, codec $f=2$), against $7.8\times10^{-3}$ with GELU on the same grids.
The codec needs $f=2$ because its fine-band output of degree $2L+1$ must
represent the quadratic product rather than its projection; at $f=1.5$ the
error rises to $2.4\times10^{-2}$. On the shallow-water task, whose codec is
linear, $f=1.5$ throughout: over five seeds the quadratic at $f=1.5$ reaches
119 and 211 ($\times10^{-3}$) at five and ten hours against 124 and 217 for
GELU at $f=3$, with the trained equivariance error of both at the
floating-point floor.

The spectral normalization $N_{\mathrm{spec}}$ in
Eq.~\eqref{eq:grid-update} centers the invariant $(\ell,m)=(0,0)$ coefficient
and normalizes by the total spectral energy. Both operations are invariant to
the action within each degree and therefore preserve equivariance.

\subsection{Encoder, decoder, and output representation}
\label{app:codec}

The encoder applies its first pointwise layer on the data grid, band-limits the
result, resamples it to the oversampled grid, and applies the activation and
second layer there before the analysis transform. The decoder concatenates the
band-limited hidden field with the band-limited raw input on the oversampled
grid and applies its two pointwise layers there. Its output is retained as a
spherical function at the fine grid's own band limit, $2L+1$ on WeatherBench,
then sampled at the data-grid nodes and combined with a pointwise linear skip
from the raw input.

The fine-band output is important at the WeatherBench resolution. On a
$32\times64$ grid, approximately half of the grid degrees of freedom lie
outside the space spanned by harmonics through degree 31. Projecting the target
onto this truncated space alone produces a normalized MSE of
$33\times10^{-3}$, which is already a substantial fraction of the forecast
error of the strongest models in Table~\ref{tab:wb}. We therefore retain the
fine-band spherical function produced by the decoder rather than truncating the
forecast to the bandwidth of the residual stream. The corresponding ablation is
reported in Appendix~\ref{app:output-residual}.

\subsection{Why low-frequency coupling is sufficient}
\label{app:band}

The final model uses the coupled band $12/24$, which keeps the output cutoff at
twice the input cutoff and therefore retains the complete correction generated
by its low-degree inputs. Widening the coupling to the full $31/31$ spectrum
under the same uniform-oversampling recipe reaches 93.5 ACC and 80.5 MSE at one
day, compared with 93.4 ACC and 80.7 MSE for $12/24$ (single seeds). The wider
band therefore produces no material forecast gain in this comparison, while
the tiled-kernel cost rises from 4.6 to 35.6\,ms per block
(Table~\ref{tab:cgblock}).

Figure~\ref{fig:spectrum} gives a data-side explanation for this result. From
the power per degree of 512 WeatherBench test states, degrees $\ell\le12$ hold
81\% of the variance of the six target fields and 83\% of the 24 input
channels, while degrees $\ell\le24$ hold 96\% of target variance and 88\% of
the one-day tendency variance. The coupled input band therefore contains most
of the large-scale atmospheric state, and the output band contains most of the
change that the forecast must produce. Degrees above the coupled band are not
discarded: they continue through the per-degree linear maps and grid MLPs in
every block. In contrast, the SSWE initial conditions are spectrally white up
to degree 85, so no low-frequency band is privileged on that task.

\begin{figure}[htbp]
  \centering
  \includegraphics[width=0.7\linewidth]{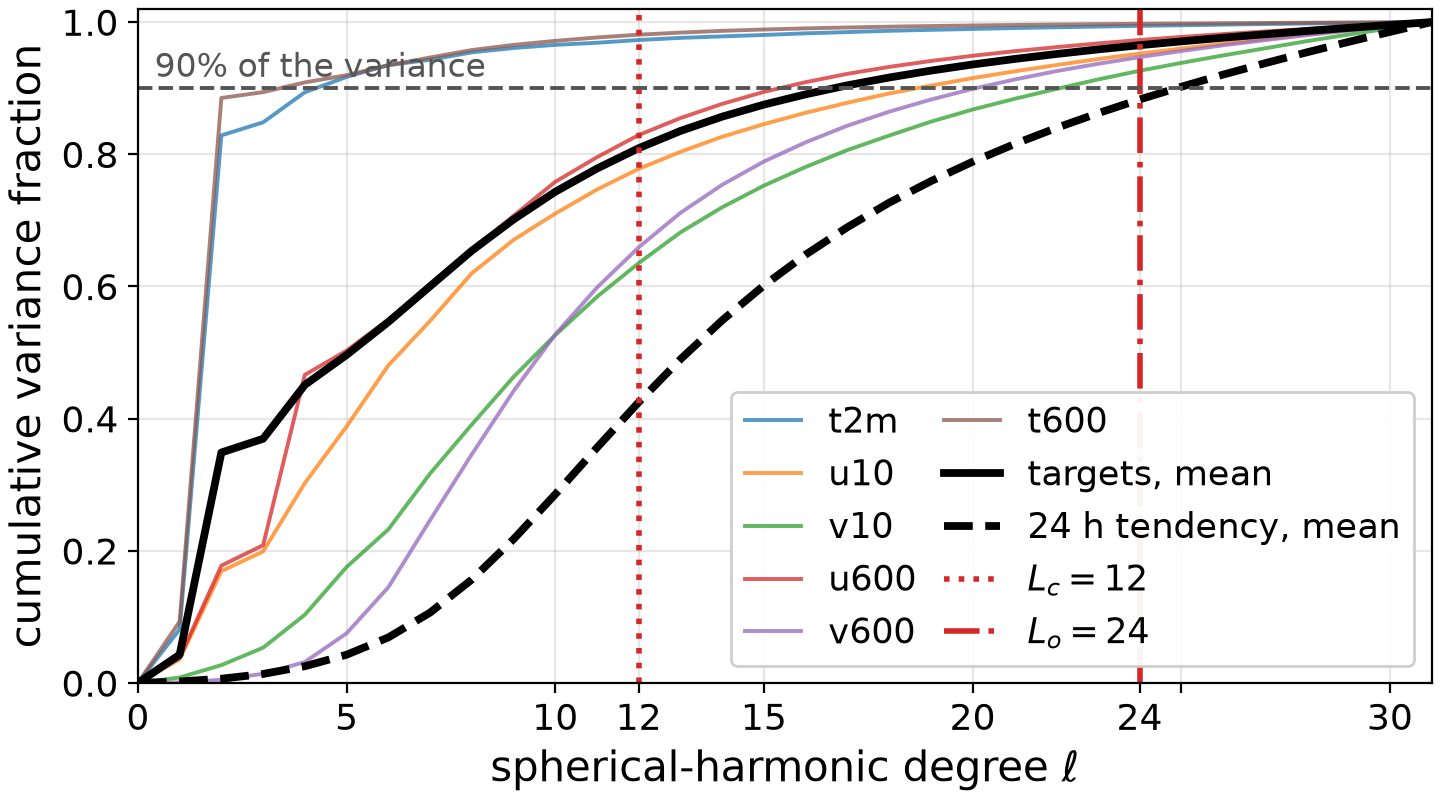}
  \caption{Cumulative fraction of WeatherBench variance below each
  spherical-harmonic degree over 512 test states. The dotted lines mark the
  input and output cutoffs of the coupled band. The input cutoff covers most of
  the state variance, while the output cutoff covers nearly all target variance
  and most one-day tendency variance.}
  \label{fig:spectrum}
\end{figure}

\fi

\ifdefined\AppendixAcceleration
\section{SSHCG: An Efficient Clebsch--Gordan Evaluation Library}
\label{app:acceleration}

The main text reports the computational benefit of SSHCG through block-level
and end-to-end timings. This section provides the implementation details and
numerical validation behind those results. We first examine the representation
costs of existing tensor-product libraries, then describe the sparse direct
kernels used by SSHCG. We next introduce the factorized Gaunt route for wider
coupled bands and report numerical agreement and model-level runtime. Finally,
we evaluate SSHCG on tensor-product workloads derived from MACE and NequIP,
as well as on broader synthetic spectral configurations.

\subsection{Representation cost of existing tensor-product libraries}
\label{app:sshcg-motivation}

\paragraph{Pathwise dense contraction in \texttt{e3nn}~\citep{geiger2022e3nn}.}
Broad-band CG coupling gathers sparse angular contributions from many degree
paths into a shared output. In a pathwise dense implementation, each degree
path is contracted over an angular domain of
$(2\ell_1+1)(2\ell_2+1)(2\ell+1)$ entries and produces a separate intermediate
result before accumulation into the final output. Two overheads therefore grow
with bandwidth, as illustrated in Figure~\ref{fig:sshcg-motivation}a,b: the
dense angular domain relative to the nonzero CG coefficients it contains, and
the logical volume of pathwise intermediate results relative to the final
output.

\paragraph{Channel expansion in \texttt{cuEquivariance}~\citep{nvidia2026cuequivariance}.}
The weighted channelwise operator considered here shares the same angular
connectivity across channels. In the tested public descriptor, however, the
supported kernel representation scalarizes angular coordinates and channels
together. Consequently, $N(L)$ nonzero angular couplings become $C N(L)$
scalar path records. Figure~\ref{fig:sshcg-motivation}c shows the resulting
linear growth of registered metadata with channel count at fixed bandwidth.

\begin{figure}[htbp]
  \centering
  \includegraphics[width=\linewidth]{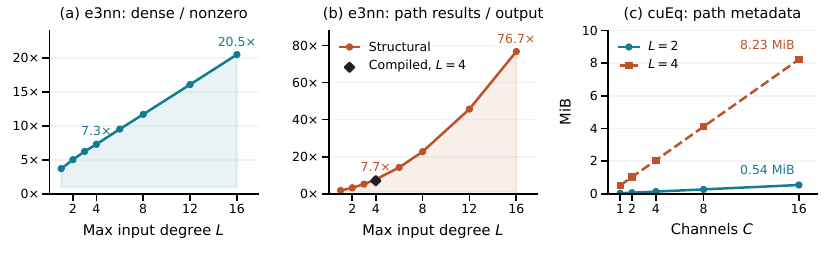}
  \caption{
  Representation costs of broad-band tensor products.
  (a) Dense \texttt{e3nn} coefficient domain relative to real-basis nonzeros.
  (b) Logical path-result volume relative to the final output; the diamond marks
  the compiled observation.
  (c) Registered path metadata in the constructed \texttt{cuEquivariance}
  operator, whose backend path count equals the channel count $C$ times the
  number of angular nonzeros in every tested configuration.
  }
  \label{fig:sshcg-motivation}
\end{figure}

\subsection{Sparse direct CG kernels}
\label{app:sshcg-direct}

The direct CG product repeatedly traverses angular domains whose dense size
grows substantially faster than the number of nonzero couplings. SSHCG instead
stores the channel-independent angular geometry once and organizes the forward
pass around output-owned tiles. Fixing $m_2=m-m_1$ restricts traversal to the
legal angular domain, while bucketed contraction widths allow each Triton
program to accumulate contributions from multiple degree paths into an output
tile before a single write.

The backward pass is organized by reduction destination. For nonzero angular
couplings $e$,
\begin{equation}
  \bar{x}_i
  =\sum_{e:\,i(e)=i}
  a_e w_{p(e)}y_{j(e)}\bar{z}_{k(e)},
  \qquad
  \bar{w}_p
  =\sum_{e:\,p(e)=p}
  a_e x_{i(e)}y_{j(e)}\bar{z}_{k(e)},
  \label{eq:sshcg-backward}
\end{equation}
so input gradients are gathered by signed spectral coordinate through a reverse
compressed-sparse-row traversal, while weight gradients are reduced over degree
paths and batches. This organization avoids materializing a separate dense
intermediate for every degree path and allows the same angular geometry to be
reused across neighboring channels. Appendix~\ref{app:sshcg-motivation}
contrasts this representation with the layouts used by existing tensor-product
libraries.

\subsection{Factorized Gaunt evaluation}
\label{app:gaunt}

The sparse direct kernels reduce the constant factor of the CG contraction, but
the number of admissible degree triples still grows rapidly with bandwidth. For
wider coupled bands, we therefore provide a second implementation that changes
the computational scaling itself. When the degree-path weights admit a low-rank
canonical polyadic (CP) factorization, the coupling can be evaluated through
spherical synthesis, a bilinear operation in physical space, and harmonic
analysis, following the Gaunt-product formulation of
\citet{luo2024gaunt} and the odd-sector Poisson construction of
\citet{bochkarev2026contracted}:
\begin{equation}
  z_{bc,\ell m}
  = \sum_s\sum_{r=1}^{R_s}D^{(s)}_{cr\ell}
  \left[\cA\,\cB_s\!\left(
  \cS\!\left[A^{(s)}_{cr\ell}x_{bc,\ell m}\right],
  \cS\!\left[B^{(s)}_{cr\ell}y_{bc,\ell m}\right]
  \right)\right]_{\ell m},
  \label{eq:gaunt}
\end{equation}
where $s$ indexes the even- and odd-parity sectors, $R_s$ is the corresponding
CP rank,
$\cB_{\mathrm{even}}(F,G)=FG$, and
\[
\cB_{\mathrm{odd}}(F,G)
=
\hat{\mathbf r}\cdot
(\nabla_{S^2}F\times\nabla_{S^2}G).
\]
Under sufficiently accurate quadrature, this construction exactly realizes the
corresponding CP-parameterized operator. Its complexity changes from the
$O(BCL^5)$ angular enumeration of broad-band direct coupling to the
$O(BCRL^3)$ transform-based evaluation. The grid route therefore becomes
increasingly favorable as the coupled bandwidth grows, provided that the CP
rank remains modest. Figure~\ref{fig:sshcg-gaunt-crossover} empirically confirms
the resulting crossover in runtime across bandwidth, batch size, precision,
and CP rank.

Under the corresponding CP-to-LCG parameter mapping, the Gaunt and sparse
implementations produce numerically consistent outputs and gradients. The
maximum relative $L_2$ errors are $1.3\times10^{-14}$ in FP64 and
$3.5\times10^{-7}$ in FP32, validating the Gaunt route as a numerically
equivalent realization of the same CP-parameterized operator.

\begin{figure}[htbp]
  \centering
  \includegraphics[width=\linewidth]{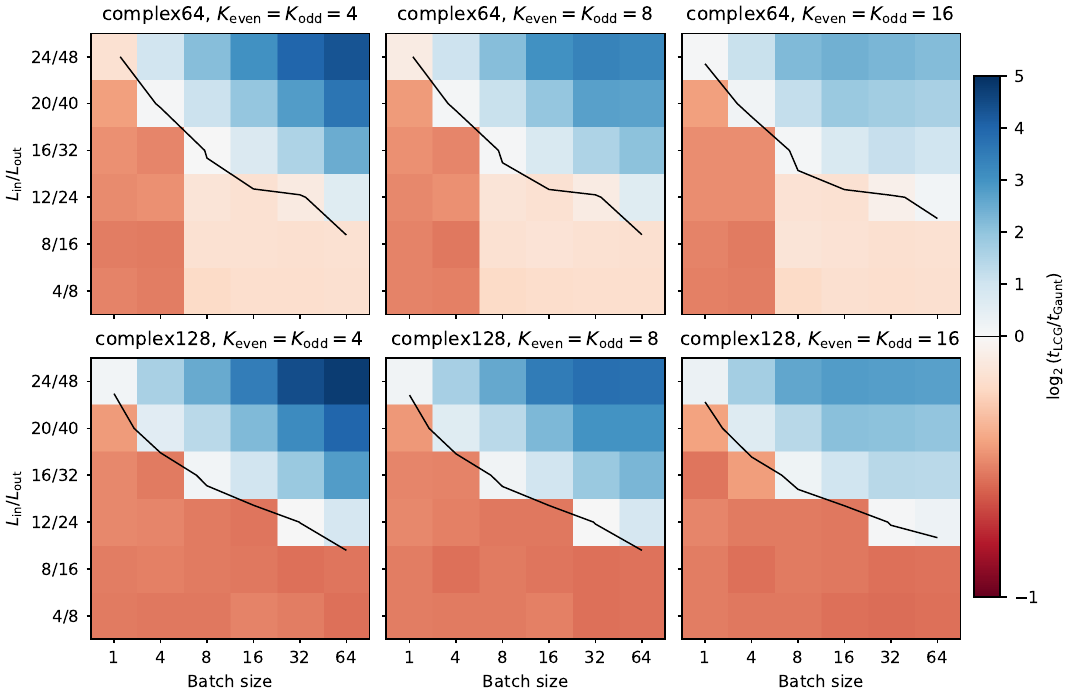}
  \caption{
  Matched-operator eager forward-plus-backward crossover as bandwidth,
  batch size, precision, and per-sector CP rank vary. Color represents
  $\log_2(t_{\mathrm{LCG}}/t_{\mathrm{Gaunt}})$; blue favors Gaunt, red favors
  direct CG, and the black contour marks equal latency. We use channel
  multiplicity eight, batch sizes $B\in\{1,4,8,16,32,64\}$, and equal even-
  and odd-sector ranks $K\in\{4,8,16\}$.
  }
  \label{fig:sshcg-gaunt-crossover}
\end{figure}

\subsection{Runtime and numerical validation}
\label{app:speed}

\paragraph{WeatherBench.}
Table~\ref{tab:cgblock} reports the forward-and-backward latency of a single CG
block, while Figure~\ref{fig:speed} reports the corresponding end-to-end model
cost. At the default $12/24$ coupled band, the CG interaction accounts for
24\% of a full training step with the dense product, 11\% with the fused
kernels, 10\% with the tiled kernels, and 9\% with the grid route. At $16/31$,
the corresponding shares are 47\%, 19\%, 14\%, and 10\%. At the full band,
where the dense product no longer fits in memory, the fused kernels account for
71\% of the step and the tiled kernels for 40\%, whereas the grid route remains
at 11\%. The grid route therefore stays near its default-band cost as the
coupling is extended toward the full spectrum; at the full band, the total
training-step time is only 2\% higher than at the default band.

The fused and tiled kernels evaluate the same direct contraction as the dense
reference and agree with it to within $2\times10^{-7}$ in the forward pass and
$5\times10^{-7}$ in the gradients in single precision. The grid route instead
uses a factorized realization of the same equivariant interaction structure.
When trained independently on the 24-hour WeatherBench task, it obtains
93.4 ACC and 81.1 MSE, compared with 93.4 ACC and 80.7 MSE for the
dense-product parameterization. Together with its weak dependence on coupled
bandwidth, this makes the grid route a practical alternative when extending the
interaction toward the full spectrum.

The main model nevertheless uses the tiled direct kernels at the operating
$12/24$ band. First, the tiled and grid implementations differ by less than a
millisecond per block (4.6 versus 4.1\,ms) and by less than one percent per
training step (269 versus 268\,ms with TF32 disabled), so the grid route offers
little benefit at this bandwidth. Second, exact reproduction of the
factorized coupling requires strict single-precision matrix multiplications in
the transforms, which also disables TF32 for the remaining layers. Under the
mixed-precision setting used by the main model, the tiled implementation
therefore requires 233\,ms per step, compared with 268\,ms for the grid-route
model. The step times of this subsection are measured on the uniform
oversampling configuration of the acceleration study, with blocks and codec
both at $f=2$; the main model runs its blocks at $f=1.5$ and takes 176\,ms
under the same setting (Appendix~\ref{app:dealiasing}). Third, the tiled kernels retain an independent learned path weight
$\gamma$ for every admissible degree triple, whereas the grid route restricts
these weights to a rank-$R$ CP factorization. Since the unrestricted
parameterization remains affordable at the operating band, we use the tiled
direct implementation there and reserve the grid route for wider coupled bands.

\paragraph{MACE- and NequIP-derived workloads.}
Although SSHCG is developed for the broad-band interactions required by our
forecasting model, its sparse direct kernels also apply to CG tensor products
used in equivariant interatomic potential models. To evaluate this setting,
Table~\ref{tab:sshcg-task} compares \texttt{e3nn}, SSHCG, and
\texttt{cuEquivariance} on operator configurations derived from MACE
\citep{batatia2022mace} and NequIP~\citep{batzner2022nequip}. The benchmarks use
per-sample weights and path-separated outputs under a common real-basis
interface. SSHCG achieves the lowest forward--backward latency in every
reported configuration.

The MACE workloads fix the feature width while varying the angular complexity
of the tensor product. The NequIP workloads vary both feature width and maximum
angular degree across five archived operator configurations. In all cases, we
benchmark the tensor-product operator in isolation rather than the complete
MACE or NequIP model, so the measurements reflect implementation-level
differences in CG execution without contributions from unrelated model
components.

Each timing uses ten warm-up iterations followed by five rounds of 30 measured
iterations, and reports the median of the round means obtained from CUDA events.
CUDA Graph capture uses separate static buffers and is validated against eager
outputs and gradients, with external pre-replay copies excluded from the
measurement. All library--configuration pairs pass the numerical checks in
their evaluated precision. The maximum relative $L^2$ errors across outputs,
input gradients, and weight gradients are $2.436\times10^{-7}$ in FP32 and
$3.084\times10^{-15}$ in FP64.

\begin{table*}[t]
\centering
\small
\setlength{\tabcolsep}{5pt}
\caption{
Task-derived UVU configurations and forward--backward latency
(CUDA-event ms), FP32, CUDA Graph, A100 PCIe 40GB, $B=1024$.
Degree ranges are inclusive.
}
\label{tab:sshcg-task}
\begin{tabular}{lccccc}
\toprule
Task
& $C_u$
& \shortstack{Degrees\\(L/R/O)}
& \shortstack{e3nn}
& \shortstack{SSHCG}
& cuEquivariance \\
\midrule
MACE-large
& 128 & 0--2 / 0--3 / 0--3
& 2.02 & \textbf{1.04} & 2.30 \\

MACE-medium
& 128 & 0--1 / 0--3 / 0--2
& 0.774 & \textbf{0.345} & 0.399 \\

NequIP-LiPS
& 32 & 0--2 / 0--2 / 0--2
& 1.58 & \textbf{0.665} & 1.32 \\

NequIP-aspirin
& 64 & 0--1 / 0--1 / 0--1
& 0.516 & \textbf{0.175} & 0.225 \\

NequIP-toluene
& 64 & 0--2 / 0--2 / 0--2
& 2.18 & \textbf{0.756} & 2.79 \\

NequIP-benzene
& 64 & 0--3 / 0--3 / 0--3
& 7.94 & \textbf{2.83} & 17.3 \\

NequIP-water
& 32 & 0--1 / 0--1 / 0--1
& 0.441 & \textbf{0.162} & 0.287 \\
\bottomrule
\end{tabular}
\end{table*}

\paragraph{Broad-band Clebsch--Gordan benchmark.}
The MACE- and NequIP-derived workloads above involve relatively low angular
degrees. To evaluate the behavior of SSHCG at wider spectral bandwidths, we
therefore construct an additional benchmark that isolates the
Clebsch--Gordan tensor-product operator.

We consider two coupling families. S1 measures broad-band interactions between
two spectra supported on $[0,L]$, with the output retained on $[0,2L]$. S2
measures cross-band interactions between a low-frequency spectrum supported on
$[0,K]$ and a higher-frequency spectrum supported on $[K+1,H]$, with outputs
retained on $[K+1,H]$. All cutoff degrees are inclusive, and every
triangle-valid degree path is included with both degree parities. The CG weights
are independent across degree paths and channels and shared across the batch.
FP32 and FP64 correspond to \texttt{complex64} and \texttt{complex128}
half-spectra, respectively. Flat spectra use amplitude
$1/\sqrt{2\ell+1}$ with a case-specific seeded CPU generator.

Outputs and vector--Jacobian products are checked against an independent FP64
reference. We use the scaled maximum error
\[
\frac{\|v-v_{\rm ref}\|_\infty}
{\max(1,\|v_{\rm ref}\|_\infty)},
\]
with tolerances $5\times10^{-4}$ in FP32 and $10^{-10}$ in FP64, together
with finite-value and dtype checks. Partial timings from failed runs are
excluded, and timeouts are not reported as steady-state latency.

\begin{table}[htbp]
\centering
\small
\setlength{\tabcolsep}{4pt}
\caption{
Spectral forward--backward latency (ms), FP32, CUDA Graph,
A100 SXM4 80GB, $B=C=8$.
}
\label{tab:sshcg-main}
\begin{tabular}{lccccc}
\toprule
Workload
& $L/H$
& \shortstack{e3nn}
& SSHCG
& \shortstack{cuEquivariance} \\
\midrule
S1 & 4  & 1.47 & \textbf{0.0738} & 7.59 \\
S1 & 8  & 7.47 & \textbf{0.226}  & 180 \\
S1 & 12 & 27.3 & \textbf{0.832}  & TO \\
S2 & 8  & 1.17 & \textbf{0.119}  & 13.6 \\
S2 & 16 & 4.55 & \textbf{0.284}  & 129 \\
S2 & 24 & 8.74 & \textbf{0.472}  & TO \\
\bottomrule
\end{tabular}

\par\smallskip
{\footnotesize
Bold indicates the row minimum.
S1: $L_o=2L$; S2: $K=4$, $L_o=H$.
TO denotes timeout.
}
\end{table}

Across the reported spectral configurations, the direct SSHCG implementation
substantially reduces tensor-product latency. For S1, its speedup over compiled
\texttt{e3nn} increases from $19.9\times$ at $L=4$ to $32.8\times$ at
$L=12$. At $L=8$, the complete forward--backward pass requires 0.226\,ms with
SSHCG, compared with 7.47\,ms for compiled \texttt{e3nn} and 180\,ms for the
evaluated \texttt{cuEquivariance} route. These measurements compare complete
operator execution paths and therefore do not attribute the observed speedup to
any single implementation decision.

\fi

\ifdefined\AppendixProtocol
\section{Experimental Protocol}
\label{app:protocol}

This section fixes the common protocol used to answer the four experimental
questions in Section~\ref{sec:experiments}. It specifies the datasets and
metrics, optimization objective, model configurations, and rotation diagnostic
before the component-level evidence is examined.

\subsection{Benchmarks and evaluation}
\label{app:benchmarks}

\paragraph{Spherical shallow-water equations.}
For the controlled spherical-dynamics benchmark, we use the
torch-harmonics solver at $256\times512$ resolution with a one-hour
model step, fresh random initial conditions in every epoch, and 50 fixed test
initial conditions. Performance is measured by per-variable relative $L^2$
rollout error in physical units at one, five and ten hours, averaged over height,
vorticity, and divergence. The solver contains spherical-harmonic content up
to degree 85. Because truncating the model at degree 63 produces a normalized
error floor of 0.49, all spectral models in our study use $L=85$ on this task.

\paragraph{WeatherBench.}
For global atmospheric forecasting, we use the WeatherBench benchmark
constructed from ERA5 reanalysis at $5.625^\circ$ ($32\times64$)
resolution~\citep{rasp2020weatherbench}. The input contains 24 prognostic
variables (2T, 10U, 10V, and U, V, T at seven pressure levels), together with
land--sea mask, orography, and latitude. The six prediction targets are 2T,
10U, 10V, U600, V600, and T600. We use hourly samples with training years
1979--2015, validation year 2016, and test years 2017--2018, and train one
direct-forecast model for each lead time of 24, 72, and 120 hours.

We report anomaly correlation coefficient (ACC) against the split climatology
and latitude-weighted MSE of normalized variables, averaged over the six
targets. WeatherBench's offset equiangular latitudes are linearly interpolated
onto Legendre--Gauss nodes so that all spherical operators are trained and
evaluated on the same grid.

\paragraph{Baselines.}
ClimaX, FourCastNet, and SHNet use the authors' implementations at the model
sizes specified by the benchmark paper. Our reference SFNO
follows the corresponding published implementation. All models in
Tables~\ref{tab:sswe}, \ref{tab:wb}, and~\ref{tab:equiv} are trained by us on the same data
splits and grids.

\subsection{Training objective and optimization}
\label{app:objective}

For $C$ prognostic variables and quadrature weights $w_{ij}$, the one-step
training loss is the spherically weighted relative $L^2$ objective used by the
benchmark,
\begin{equation}
  \mathcal{L}
  =\frac{1}{C}\sum_{q=1}^{C}
  \left(
  \frac{\sum_{i,j}w_{ij}(\widehat{x}_{qij}-x_{qij})^2}
       {\sum_{i,j}w_{ij}x_{qij}^2}
  \right)^{1/2}.
  \label{eq:one-step-loss}
\end{equation}
The loss is computed in normalized variables, and no rollout fine-tuning is
used.

For WeatherBench, optimization follows the benchmark protocol: AdamW with
learning rate $4\times10^{-4}$, weight decay $10^{-5}$, batch size 128, linear
warm-up, cosine decay, and a maximum of 50 epochs with early stopping on
validation RMSE using patience 3. For the spherical shallow-water task, we use
Adam with learning rate $10^{-3}$, batch size 16, and 50 epochs of 256 fresh
samples per epoch.

\subsection{Model configurations and hyperparameters}
\label{app:hyper}

\begin{table}[htbp]
  \caption{Architectures of the compared models at the sizes of the benchmark paper, and the training length of our models (epochs run until early stopping at 1, 3 and 5 days on WeatherBench; fixed budget on the shallow-water task). WeatherBench models use the benchmark optimiser and schedule (AdamW, learning rate $4\times10^{-4}$, batch 128, warm-up and cosine decay, early-stopping patience 3); shallow-water models use the benchmark recipe (Adam, learning rate $10^{-3}$, batch 16, 50 epochs).}
  \label{tab:hparams}
  \centering
  \footnotesize
  \setlength{\tabcolsep}{4pt}
  \begin{tabular}{llr@{\hspace{6pt}}p{0.50\textwidth}}
    \toprule
    Benchmark & Model & Params & Architecture (epochs run) \\
    \midrule
    Shallow water & ClimaX & 1.96M & patch 4, width 128, depth 6, 8 heads \\
    & FourCastNet & 1.66M & patch 4, width 128, depth 8, 8 AFNO blocks \\
    & SHNet (GSNO) & 0.87M & width 8, scale factor 4 \\
    & SFNO & 0.28M & width 28, 4 blocks, $L=85$, scale factor 2 \\
    & LCG-SFNO & 0.27M & width 25, 4 blocks, $L=85$, CG band 12/24 rank 4, quadratic activation, oversampling 1.5 \\
    \midrule
    WeatherBench & ClimaX & 5.40M & patch 4, width 256, depth 6, 8 heads, decoder depth 2 \\
    & FourCastNet & 5.60M & patch 4, width 384, depth 8, 8 AFNO blocks \\
    & SHNet (GSNO) & 4.03M & U-shaped, width 64, scale factor 2, encoder layers 2 \\
    & SFNO & 5.26M & width 256, 4 blocks, $L=15$, scale factor 2 \\
    & LCG-SFNO & 5.24M & width 186, 4 blocks, $L=31$, CG band 12/24 rank 4, quadratic activation, oversampling 1.5 in the blocks and 2.0 in the codec, two-layer codec, drop path 0.1, dropout 0.1 (43, 11, 5) \\
    \bottomrule
  \end{tabular}
\end{table}

\subsection{Equivariance evaluation protocol}
\label{app:equivariance}

For a trained model $\cF$ and a rotation $R$ drawn uniformly from $\SO(3)$, we
report
\begin{equation}
\epsilon_{\mathrm{eq}}
=
\frac{\left\|\cF(R\bx,R\bc)-R\,\cF(\bx,\bc)\right\|_2}
{\left\|R\,\cF(\bx,\bc)\right\|_2},
\label{eq:equivariance-diagnostic}
\end{equation}
averaged over eight rotations and sixteen test inputs using the area-weighted
norm on the Legendre--Gauss grid. Rotations are sampled uniformly from
$\SO(3)$ using ZYZ Euler angles and are applied to real spherical-harmonic
coefficients degree by degree with Wigner $D$-matrices obtained from the matrix
exponential of the angular-momentum generator; no spatial interpolation is
used.

Inputs are band-limited before rotation, and all channels, including static
conditioning fields, are rotated jointly. Projection onto degrees $\ell\le L$
commutes with rotation, so band-limiting the input before rotation and the
output before comparison does not alter the equivariance relation being tested.
For models whose decoder emits a spherical function with bandwidth above that
of the data grid, the comparison is performed on the oversampled grid where
that function is exactly represented. All equivariance measurements are
computed with TF32 disabled. Figure~\ref{fig:equivmaps} uses the same
construction for a single test state and rotation, with both outputs projected
to the data-grid band before visualizing the residual.

\fi

\ifdefined\AppendixResults
\section{Additional Forecasting Results}
\label{app:results}

This appendix reports results omitted from the main text for space: the
variable-wise SSWE and WeatherBench scores, qualitative forecast fields, and transfer of the proposed block to a second spherical
architecture.

\subsection{SSWE rollouts and per-variable results}
\label{app:pervar-sswe}

Figure~\ref{fig:sswe-rollout} shows the relative $L^2$ error of every model over twenty one-hour steps, and Table~\ref{tab:pervar-sswe} breaks the SSWE scores of Table~\ref{tab:sswe} down by variable. The two spherical spectral models accumulate error several times more slowly than the position-dependent architectures from the first steps on; they are the most accurate on every variable at five hours, with the largest margins on divergence, where they lead at every horizon. LCG-SFNO matches or improves on SFNO step for step and on each variable.

\begin{figure}[htbp]
  \centering
  \includegraphics[width=0.7\linewidth]{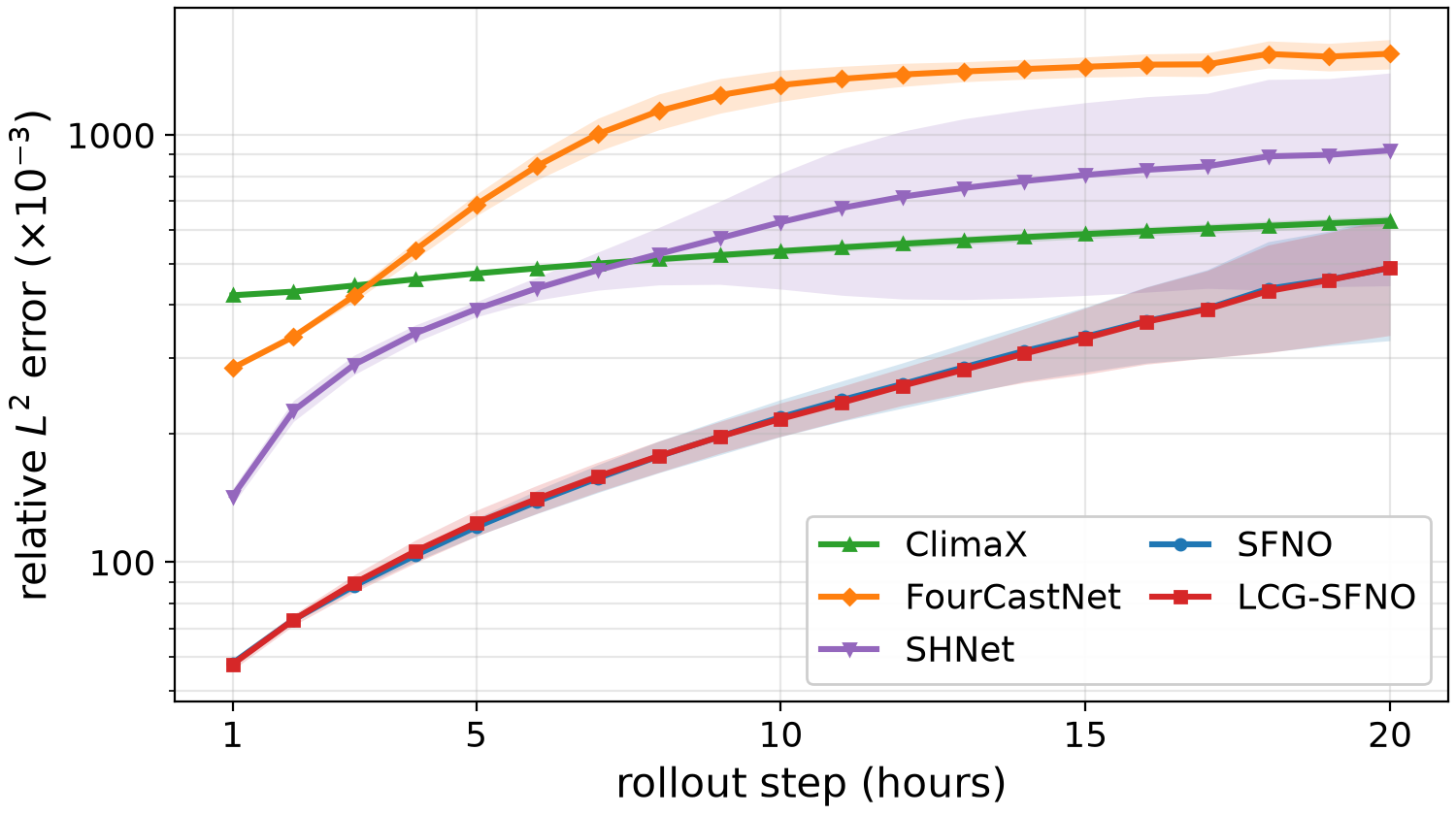}
  \caption{SSWE relative $L^2$ rollout error over twenty one-hour steps for all compared models, mean and standard deviation over five seeds with fifty test initial conditions each. The spherical spectral models accumulate error far more slowly than the position-dependent architectures, and LCG-SFNO coincides with SFNO throughout.}
  \label{fig:sswe-rollout}
\end{figure}

\begin{table}[htbp]
  \caption{SSWE forecasting per variable: relative $L^2$ rollout error in physical units ($\times10^{-3}$) of height $h$, vorticity, and divergence at one, five, and ten hours over the fifty test initial conditions, mean and standard deviation over five seeds. The best value in each column is shown in bold and the second best underlined.}
  \label{tab:pervar-sswe}
  \centering
  \scriptsize
  \setlength{\tabcolsep}{2.5pt}
  \begin{tabular}{lccccccccc}
    \toprule
    & \multicolumn{3}{c}{1 h} & \multicolumn{3}{c}{5 h} & \multicolumn{3}{c}{10 h} \\
    \cmidrule(lr){2-4}\cmidrule(lr){5-7}\cmidrule(lr){8-10}
    Model & $h$ & vorticity & divergence & $h$ & vorticity & divergence & $h$ & vorticity & divergence \\
    \midrule
    ClimaX & 5.2\,$\pm$\,0.0 & 259\,$\pm$\,1.0 & 1000\,$\pm$\,0.0 & 4.4\,$\pm$\,0.0 & 417\,$\pm$\,15 & 1000\,$\pm$\,0.1 & \textbf{3.8\,$\pm$\,0.0} & 600\,$\pm$\,31 & 1000\,$\pm$\,0.2 \\
    FourCastNet & 3.0\,$\pm$\,0.0 & \textbf{101\,$\pm$\,3.1} & 750\,$\pm$\,4.0 & 6.3\,$\pm$\,0.8 & 635\,$\pm$\,70 & 1418\,$\pm$\,125 & 6.9\,$\pm$\,1.1 & 1927\,$\pm$\,398 & 1992\,$\pm$\,245 \\
    SHNet (GSNO) & 1.4\,$\pm$\,0.1 & 147\,$\pm$\,6.4 & 279\,$\pm$\,16 & 3.9\,$\pm$\,0.8 & 419\,$\pm$\,46 & 750\,$\pm$\,19 & 7.5\,$\pm$\,7.2 & 783\,$\pm$\,314 & 1085\,$\pm$\,250 \\
    SFNO & \second{0.6\,$\pm$\,0.0} & 116\,$\pm$\,1.4 & \second{58\,$\pm$\,4.2} & \second{2.5\,$\pm$\,0.2} & \second{191\,$\pm$\,1.8} & \second{169\,$\pm$\,16} & 4.7\,$\pm$\,1.7 & \second{311\,$\pm$\,17} & \second{340\,$\pm$\,49} \\
    \textbf{LCG-SFNO (ours)} & \textbf{0.6\,$\pm$\,0.0} & \second{115\,$\pm$\,0.5} & \textbf{57\,$\pm$\,4.1} & \textbf{2.5\,$\pm$\,0.2} & \textbf{189\,$\pm$\,3.1} & \textbf{166\,$\pm$\,18} & \second{4.5\,$\pm$\,2.1} & \textbf{289\,$\pm$\,6.8} & \textbf{338\,$\pm$\,56} \\
    \bottomrule
  \end{tabular}
\end{table}

\subsection{Per-variable WeatherBench results}
\label{app:pervar}

Tables~\ref{tab:pervar-acc} and~\ref{tab:pervar-mse} give the variable-wise
breakdown of the aggregate WeatherBench results in Table~\ref{tab:wb}. At one day the improvement over
SFNO holds for all six reported variables, with the largest gains on the wind
components and the highest ACC on both temperature fields. At three and five days, the leading models move closer together and the relative ordering changes across methods. LCG-SFNO nevertheless retains a substantial aggregate improvement over the reference SFNO, with the largest gains continuing to appear in the wind variables.

\begin{table}[htbp]
  \caption{WeatherBench $5.625^\circ$ per variable, anomaly correlation coefficient in percent (higher is better), by lead time. All models trained by us; mean over three seeds. The best value in each column and block is in bold and the second best underlined.}
  \label{tab:pervar-acc}
  \centering
  \small
  \setlength{\tabcolsep}{4pt}
  \begin{tabular}{lccccccc}
    \toprule
    Model & 2T & 10U & 10V & U600 & V600 & T600 & Average \\
    \midrule
    \multicolumn{8}{l}{\emph{One day}} \\
    ClimaX & 94.3 & \textbf{91.4} & \textbf{91.5} & \textbf{93.3} & \textbf{92.2} & \second{96.7} & \second{93.2} \\
    FourCastNet & 94.9 & \second{90.8} & \second{90.8} & 92.6 & 91.3 & 96.5 & 92.8 \\
    SHNet (GSNO) & \second{97.1} & 89.5 & 89.4 & 91.7 & 90.2 & 96.5 & 92.4 \\
    SFNO & 95.3 & 76.5 & 74.9 & 83.8 & 80.1 & 93.3 & 84.0 \\
    \textbf{LCG-SFNO} & \textbf{97.4} & 90.6 & 90.3 & \second{92.9} & \second{91.6} & \textbf{97.1} & \textbf{93.3} \\
    \midrule
    \multicolumn{8}{l}{\emph{Three days}} \\
    ClimaX & 90.6 & \textbf{71.1} & \textbf{69.4} & \textbf{78.2} & \textbf{71.4} & \textbf{88.8} & \textbf{78.3} \\
    FourCastNet & 89.9 & 61.0 & 55.6 & 69.3 & 56.6 & 83.4 & 69.3 \\
    SHNet (GSNO) & \textbf{92.9} & \second{65.3} & \second{62.9} & \second{72.9} & \second{64.7} & \second{86.4} & \second{74.2} \\
    SFNO & 89.5 & 52.4 & 49.5 & 64.6 & 54.6 & 81.6 & 65.4 \\
    \textbf{LCG-SFNO} & \second{92.7} & 64.2 & 61.6 & 72.7 & 64.0 & 86.1 & 73.5 \\
    \midrule
    \multicolumn{8}{l}{\emph{Five days}} \\
    ClimaX & 85.9 & \second{42.3} & \second{34.8} & \second{50.6} & \second{31.3} & \second{73.7} & \second{53.1} \\
    FourCastNet & 87.0 & 40.0 & 31.4 & 47.9 & 25.6 & 71.3 & 50.5 \\
    SHNet (GSNO) & \textbf{89.1} & \textbf{43.6} & \textbf{36.7} & \textbf{52.1} & \textbf{33.3} & \textbf{74.5} & \textbf{54.9} \\
    SFNO & 85.9 & 34.0 & 28.1 & 45.5 & 26.3 & 71.8 & 48.6 \\
    \textbf{LCG-SFNO} & \second{87.1} & 40.2 & 33.5 & 48.8 & 29.8 & 70.8 & 51.7 \\
    \bottomrule
  \end{tabular}
\end{table}

\begin{table}[htbp]
  \caption{WeatherBench $5.625^\circ$ per variable, latitude-weighted MSE of normalised variables ($\times10^{-3}$, lower is better), by lead time. All models trained by us; mean over three seeds. The best value in each column and block is in bold and the second best underlined.}
  \label{tab:pervar-mse}
  \centering
  \small
  \setlength{\tabcolsep}{4pt}
  \begin{tabular}{lccccccc}
    \toprule
    Model & 2T & 10U & 10V & U600 & V600 & T600 & Average \\
    \midrule
    \multicolumn{8}{l}{\emph{One day}} \\
    ClimaX & 11.8 & \textbf{89.9} & \textbf{136} & \textbf{68.5} & \textbf{144} & \second{9.87} & \textbf{76.8} \\
    FourCastNet & 9.66 & \second{96.0} & \second{146} & 75.3 & 160 & 10.5 & 82.8 \\
    SHNet (GSNO) & \second{6.21} & 109 & 167 & 84.5 & 179 & 10.7 & 92.7 \\
    SFNO & 9.93 & 228 & 367 & 158 & 344 & 19.8 & 188 \\
    \textbf{LCG-SFNO} & \textbf{5.55} & 97.3 & 154 & \second{72.1} & \second{155} & \textbf{8.86} & \second{82.1} \\
    \midrule
    \multicolumn{8}{l}{\emph{Three days}} \\
    ClimaX & 19.1 & \textbf{274} & \textbf{440} & \textbf{208} & \textbf{480} & \textbf{32.3} & \textbf{242} \\
    FourCastNet & 20.0 & 348 & 589 & 277 & 670 & 46.0 & 325 \\
    SHNet (GSNO) & \textbf{14.9} & \second{316} & \second{507} & \second{249} & \second{564} & \second{38.7} & \second{282} \\
    SFNO & 21.6 & 403 & 639 & 310 & 682 & 49.9 & 351 \\
    \textbf{LCG-SFNO} & \second{15.4} & 323 & 521 & 250 & 571 & 39.3 & 287 \\
    \midrule
    \multicolumn{8}{l}{\emph{Five days}} \\
    ClimaX & 28.6 & 459 & 753 & \second{400} & 901 & \second{67.6} & 435 \\
    FourCastNet & 26.1 & 464 & 757 & 410 & 912 & 71.7 & 440 \\
    SHNet (GSNO) & \textbf{22.4} & \textbf{446} & \textbf{729} & \textbf{387} & \textbf{871} & \textbf{65.7} & \textbf{420} \\
    SFNO & 28.6 & 494 & 783 & 421 & 910 & 70.5 & 451 \\
    \textbf{LCG-SFNO} & \second{25.7} & \second{459} & \second{742} & 402 & \second{883} & 71.9 & \second{430} \\
    \bottomrule
  \end{tabular}
\end{table}

\subsection{Transfer to another spherical architecture}
\label{app:transfer}

To test whether the proposed operator is specific to the SFNO shell, we replace
the GSNO operator inside the U-shaped SHNet architecture with the LCG-SFNO
block. The resulting 3.85M-parameter model obtains ACC values of 92.0, 73.2,
and 52.8 at one, three, and five days, respectively. Its equivariance error is
$1.0\times10^{-3}$, compared with 0.56 for the original 4.03M-parameter
SHNet. These results indicate that the proposed block can be incorporated into
a second multiscale spherical architecture without relying on the original
SFNO network layout.

\subsection{Qualitative WeatherBench forecasts}
\label{app:qualitative}

Figures~\ref{fig:fields} and~\ref{fig:errors} compare the 24-hour forecasts of
every model of the main comparison with the target fields for one test date,
with the forecasts and the errors of all models drawn on the same colour scale
per variable. All models reproduce the large-scale temperature and circulation
structure. The error maps separate them: the position-dependent architectures
leave coherent large-scale residuals, the published SFNO configuration leaves particularly large errors in the wind
and upper-air temperature fields, while LCG-SFNO produces the smallest errors
among the spherical models on this example.

\begin{figure}[t]
  \centering
  \includegraphics[width=\linewidth]{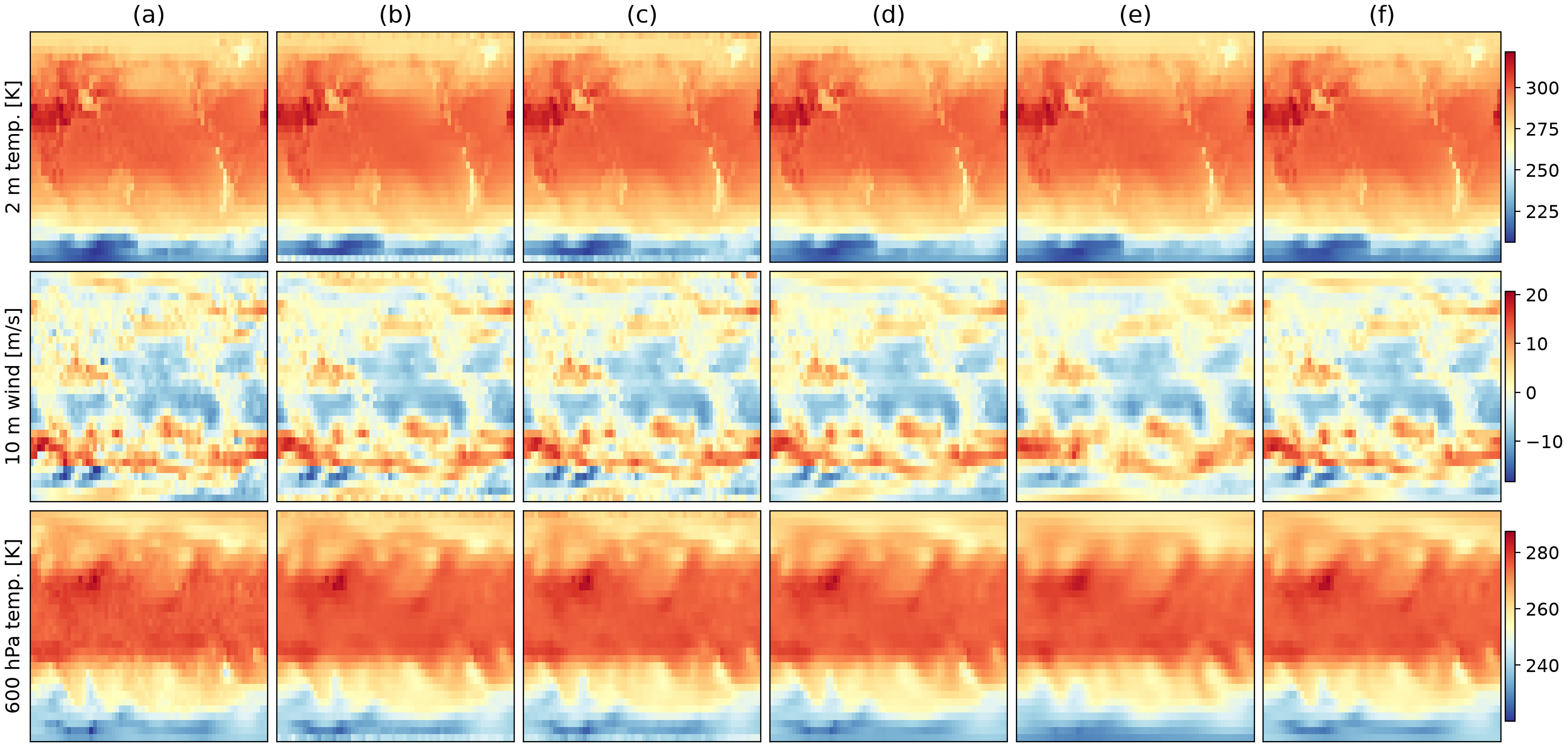}
  \caption{
  24-hour forecasts on WeatherBench $5.625^\circ$ for 15 July 2017 at 12 UTC.
  Columns (a) to (f) show the target, ClimaX, FourCastNet, SHNet, SFNO, and
  LCG-SFNO, with one colour scale per row. Rows show 2 m temperature, 10 m
  zonal wind, and 600 hPa temperature. Every model captures the large-scale
  structure; the differences lie in the finer features that the error maps of
  Figure~\ref{fig:errors} isolate.
  }
  \label{fig:fields}
\end{figure}

\begin{figure}[b]
  \centering
  \includegraphics[width=\linewidth]{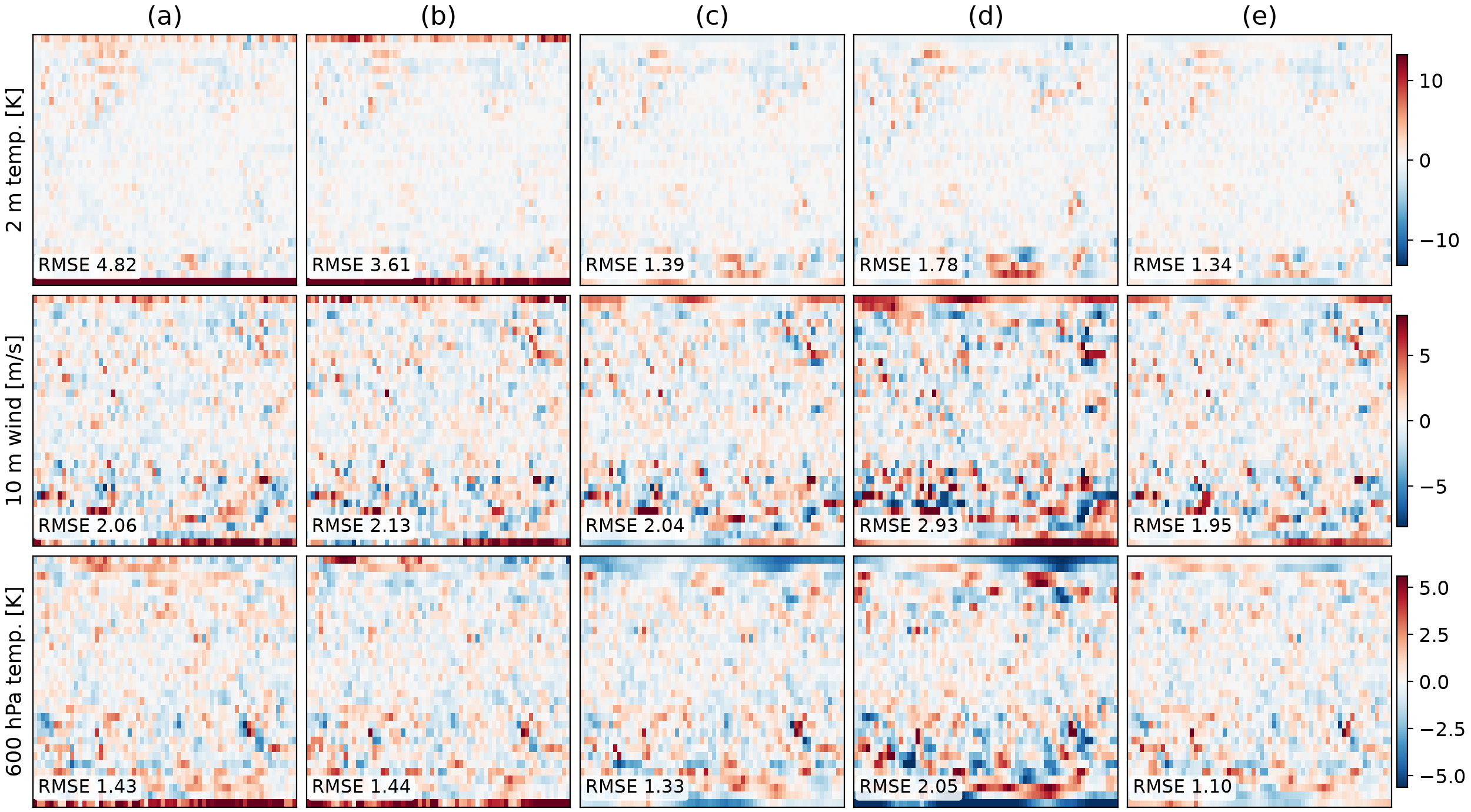}
  \caption{
  Forecast errors (forecast minus target) of the models in
  Figure~\ref{fig:fields} for the same date. Columns (a) to (e) show ClimaX,
  FourCastNet, SHNet, SFNO, and LCG-SFNO on one symmetric colour scale per
  row, set by the pooled errors of all models; each panel states the
  root-mean-square error of that model and variable on this sample. LCG-SFNO
  leaves the smallest and least coherent residuals of the spherical models.
  }
  \label{fig:errors}
\end{figure}

\fi

\ifdefined\AppendixAblations
\section{Component-Level Validation and Ablations}
\label{app:ablations}

This section provides the component-level evidence behind the main-text
results. We first separate the effects of the stronger spherical backbone, CG
coupling, and dealiased nonlinear pathway. We then justify the coupled band and
oversampling allocation, and finally test the decoder, residual pathway, and
use of positional information. The aim is to associate each architectural
choice with the function it is intended to provide rather than treating the
final model as a single indivisible change.

\subsection{Incremental architecture ablation}
\label{app:ladder}

\begin{table}[htbp]
  \caption{
  WeatherBench 24-hour ablation from the reference SFNO to LCG-SFNO.
  ACC and normalized MSE measure forecast skill; ``Epochs'' is the number of
  epochs required to reach the best validation error of row B; equivariance is
  the relative error defined in Appendix~\ref{app:equivariance}. Rows A--C are
  mean and standard deviation over three seeds; row D is a single seed. Row D
  uses the uniform oversampling configuration as a diagnostic ablation; the
  final block/codec allocation is selected in Appendix~\ref{app:oversampling}.
  All rows use the benchmark-scale parameter budget and the GELU activation;
  the final model replaces it by the quadratic activation
  (Appendix~\ref{app:dealiasing}).
  }
  \label{tab:ladder}
  \centering
  \small
  \setlength{\tabcolsep}{4pt}
  \begin{tabular}{l@{\hspace{4pt}}p{0.42\textwidth}cccc}
    \toprule
    & Model & ACC & MSE & Epochs & Equivariance \\
    \midrule
    A & SFNO, benchmark configuration
      & 84.0 $\pm$ 0.0 & 187.7 $\pm$ 0.2 & never & $9.5\times10^{-2}$ \\
    B & + full spectral band, two-layer codec, regularization
      & 93.4 $\pm$ 0.1 & 81.5 $\pm$ 0.6 & 39 & $6.7\times10^{-2}$ \\
    C & + CG coupling, activations on the data grid
      & 93.7 $\pm$ 0.0 & 77.9 $\pm$ 0.1 & 13 & $6.8\times10^{-2}$ \\
    D & + dealiased grid, band-limited residual stream
      & 93.4 & 80.7 & 21 & $7.2\times10^{-3}$ \\
    \bottomrule
  \end{tabular}
\end{table}

Table~\ref{tab:ladder} isolates the main architectural changes on 24-hour
WeatherBench forecasting. Row A is the published SFNO benchmark
configuration. Row B is a matched-capacity control that incorporates the
stronger full-band backbone, two-layer codec, input skip, stochastic depth, and
dropout, but removes the CG coupling and evaluates all nonlinearities on the
data grid. Row C adds the low-rank CG interaction while leaving the nonlinear
evaluation unchanged. Row D additionally introduces dealiased nonlinear
evaluation and the band-limited coefficient-space residual pathway.

The A$\rightarrow$B transition shows that a substantial fraction of the
forecast improvement comes from the stronger spherical backbone itself. The
B$\rightarrow$C transition isolates the CG interaction: at matched capacity it
further improves forecast error and, more prominently, reduces the number of
epochs required to reach the control's best validation error from 39 to 13.
Finally, C$\rightarrow$D primarily affects rotational consistency, reducing
the end-to-end equivariance error by nearly an order of magnitude while
retaining similar forecast skill.

The ablation therefore separates three complementary effects: the full-band
backbone provides forecast capacity, CG coupling improves the learned spectral
interaction and optimization, and dealiased nonlinear evaluation converts the
equivariant spectral structure into a substantially more equivariant
end-to-end network.

\paragraph{Why CG coupling alone does not determine end-to-end equivariance.}
The CG interaction and per-degree spectral maps are exactly equivariant by
construction. The complete network, however, also contains discretized
pointwise nonlinearities. A field band-limited at degree $L$ generates higher
spherical frequencies after the nonlinearity; on the original grid these
unresolved frequencies fold back into the retained band and create
orientation-dependent aliasing. Rows B and C therefore have similar
end-to-end equivariance errors even though row C contains the exact CG
interaction. Once the activation path is dealiased, the equivariant structure
of the spectral pathway becomes visible at the network level.

\subsection{Dealiasing and oversampling allocation}
\label{app:oversampling}

The measurements in this subsection use the GELU activation, that is, they
characterize the residual that the polynomial activation of the main model
removes, and show how that residual behaves when a non-polynomial activation is
kept.

\paragraph{Source of the end-to-end equivariance improvement.}
Rows C and D differ in the activation grid, residual stream, normalization, and
codec. Activating these changes sequentially yields initialization-time
relative equivariance errors of $2.7\times10^{-1}$ for row C,
$2.4\times10^{-1}$ with oversampled block activations,
$1.8\times10^{-1}$ after adding the band-limited residual stream and spectral
normalization, and $6.7\times10^{-3}$ once the codec is also evaluated on the
oversampled grid. The codec provides the largest single reduction in this
sequence, indicating that controlling aliasing at the input and output of the
spectral stack is especially important. Block-level oversampling provides a
further control on the residual that remains once the codec is resolved.

\paragraph{Effect of the oversampling factor.}
Varying one factor $f$ for the blocks and the codec together, the
initialization-time relative equivariance error is
$2.5\times10^{-2}$ at $f=1.5$, $6.7\times10^{-3}$ at $f=2$,
$2.6\times10^{-3}$ at $f=2.5$, and $1.0\times10^{-3}$ at $f=3$.
Over this range the error falls roughly with the square of the number of grid
points, that is, with about the fourth power of $f$, while the transform cost
grows linearly with the number of grid points. The two-factor measurements of
Table~\ref{tab:twofactor} attribute most of this reduction to the codec: at a
fixed codec factor of $2$, raising the block factor from $1.5$ to $2$ lowers the
error only from $7.8\times10^{-3}$ to $6.7\times10^{-3}$, whereas at a fixed
block factor of $1.5$ raising the codec factor from $1.5$ to $2$ lowers it from
$2.5\times10^{-2}$ to $7.8\times10^{-3}$, consistent with the sequential
activation above. The classical dealiasing rule,
$2n_\theta'\ge 3L+1$ and $n_\varphi'>3L$, is already satisfied at $f=1.5$
and controls the quadratic part of the nonlinearity, whereas the higher-order
components of GELU require additional oversampling to further suppress the
residual error.

The four internal blocks account for roughly seven tenths of the step-time
increase over row B, whereas the codec contributes less than one tenth. This
difference motivates allocating separate oversampling factors to the repeated
residual blocks and to the encoder/decoder rather than using a uniform factor.

\paragraph{Allocating oversampling between the blocks and codec.}
Because the encoder and decoder are evaluated less frequently than the internal
blocks, increasing the codec resolution is the more economical way to reduce
the residual error. Table~\ref{tab:twofactor} measures the trade-off directly.
With the blocks fixed at $f=1.5$, increasing the codec factor from 1.5 to 2.0
reduces the initialization-time equivariance error from $2.5\times10^{-2}$ to
$7.8\times10^{-3}$ while increasing the relative step time from
$2.57\times$ to $2.81\times$. A uniform factor of two reaches a similar
equivariance level at a substantially higher step time.

The GELU model of the ablations therefore keeps the blocks at the minimum factor
used by the dealiasing rule and allocates the remaining budget to the codec.
After training, this split reaches 93.4 ACC and 80.8 MSE over three seeds,
compared with 93.4 ACC and 80.7 MSE for the uniform factor of two in row D of
Table~\ref{tab:ladder}, while requiring approximately three quarters of the
time per epoch.

\begin{table}[htbp]
  \caption{
  Initialization-time equivariance error and training-step cost for different
  oversampling factors in the residual blocks and codec. Step time is
  normalized by the same architecture without CG coupling or oversampling.
  The bold row is the configuration used by the final WeatherBench model.
  }
  \label{tab:twofactor}
  \centering
  \small
  \setlength{\tabcolsep}{6pt}
  \begin{tabular}{cccc}
    \toprule
    Blocks & Codec & Equivariance & Step \\
    \midrule
    $1.5$ & $1.5$ & $2.5\times10^{-2}$ & $2.57\times$ \\
    $\mathbf{1.5}$ & $\mathbf{2.0}$ & $\mathbf{7.8\times10^{-3}}$ & $\mathbf{2.81\times}$ \\
    $1.5$ & $3.0$ & $4.1\times10^{-3}$ & $3.46\times$ \\
    $1.5$ & $4.0$ & $3.9\times10^{-3}$ & $4.30\times$ \\
    $2.0$ & $2.0$ & $6.7\times10^{-3}$ & $3.66\times$ \\
    $2.0$ & $3.0$ & $1.1\times10^{-3}$ & $4.33\times$ \\
    $2.0$ & $4.0$ & $4.7\times10^{-4}$ & $5.18\times$ \\
    \bottomrule
  \end{tabular}
\end{table}

\subsection{Output bandwidth and coefficient-space residual pathway}
\label{app:output-residual}

\paragraph{Fine-band decoding preserves forecast detail.}
A strictly degree-31 output reaches 92.2 ACC at 24 hours, whereas sampling the
fine-band spherical function produced by the decoder raises ACC to 93.4. On a
$32\times64$ grid, a substantial fraction of the grid degrees of freedom lies
above degree 31, so retaining these output degrees is important for reproducing
fine spatial structure. The final decoder therefore recovers 1.2 ACC points
while keeping the learned spectral operations equivariant.

\paragraph{Coefficient-space residuals improve rollout stability.}
The projected-residual design has little effect on one-step accuracy but
substantially improves autoregressive behavior. On SSWE with the GELU
activation at $f=3$, the earlier block design, in which the linear and CG paths
bypass the grid nonlinearity through an inner identity skip, reaches a
five-hour rollout error of 228, whereas the block without the skip, at a
comparable parameter count, reaches 115 for the same seed (124 over the five
seeds of Appendix~\ref{app:dealiasing}; the quadratic main model of
Table~\ref{tab:sswe} reaches 119). Keeping the residual stream in coefficient space and routing
new spectral content through the projected nonlinear path therefore reduces
error accumulation under repeated application of the learned operator.

\subsection{Control for positional information}
\label{app:position}

Adding a learned per-pixel positional embedding to the reference SFNO leaves
the 24-hour ACC unchanged at 84.0. This control suggests that the improvement
of LCG-SFNO is not explained by additional absolute-position capacity, but by
the changes to its spherical representation, structured spectral interaction,
and nonlinear evaluation.

\fi

\endgroup

\begingroup
\def\AppendixArchitecture{1}

\endgroup

\begingroup
\def\AppendixAcceleration{1}

\endgroup

\begingroup
\def\AppendixProtocol{1}

\endgroup

\begingroup
\def\AppendixAblations{1}

\endgroup

\begingroup
\def\AppendixResults{1}

\endgroup

\end{document}